\documentclass[10pt,twocolumn,letterpaper]{article}

\PassOptionsToPackage{pagebackref,breaklinks,colorlinks}{hyperref}
\usepackage[T1]{fontenc}
\usepackage[pagenumbers]{cvpr}

\usepackage{amsmath}
\usepackage{amssymb}
\usepackage{amsthm}
\usepackage{graphicx}
\usepackage{xcolor}
\usepackage{booktabs}

\usepackage{array}
\usepackage{multirow}
\usepackage{mathtools}
\usepackage{placeins}
\usepackage[pagebackref]{hyperref}
\newcommand{\method}{LLM-DD}

\newcommand{\figslot}[3]{%
  \IfFileExists{#1}{\includegraphics[width=\linewidth]{#1}}{%
    \fbox{\begin{minipage}[c][#2][c]{0.94\linewidth}
      \centering\small #3
    \end{minipage}}}}

\providecommand{\qwenname}{Qwen}
\providecommand{\deepseekname}{DeepSeek}
\providecommand{\pmstd}[1]{{\scriptsize\textcolor{gray}{$\pm$#1}}}

\theoremstyle{plain}
\newtheorem{llmddprop}{Proposition}

\definecolor{cvprblue}{rgb}{0.21,0.49,0.74}
\usepackage{hyperref}
\hypersetup{
    colorlinks=true,
    allcolors=cvprblue,
    pdftitle={Scaling Dataset Distillation to Large Vision-Language Models with Language-Label Mixing},
    pdfauthor={Bincheng Peng, Guang Li, Ping Liu, Takahiro Ogawa, Miki Haseyama}
}

\usepackage{algorithm}
\usepackage{algpseudocode}

\title{Scaling Dataset Distillation to Large Vision--Language Models \\ with Language--Label Mixing}

\author{%
  Bincheng Peng$^{1\dagger}$\quad Guang Li$^{1\dagger *}$\quad Ping Liu$^{2}$\quad
  Takahiro Ogawa$^{1}$\quad Miki Haseyama$^{1}$\\[4pt]
  {\normalsize $^{1}$Hokkaido University\qquad $^{2}$University of Nevada, Reno}\\[3pt]
  {\small $^{\dagger}$Equal contribution.\qquad
   $^{*}$Correspondence to Guang Li (\href{mailto:guang@lmd.ist.hokudai.ac.jp}{\texttt{guang@lmd.ist.hokudai.ac.jp}}).}}

\date{}

\begin{document}
\maketitle
\begin{abstract}
Dataset distillation for pre-trained models seeks a small synthetic set on which a lightweight head fitted on a frozen backbone matches a head fitted on the full training set. Existing methods use one-hot labels, which ignore how classes relate, and remain well below full-data accuracy at one image per class. We distill through large vision--language models (VLMs) such as Qwen3-VL and DeepSeek-VL2, whose language models supply class semantics but which provide neither a global image embedding nor a text embedding aligned with visual features. Our method, Language--Label Mixing Dataset Distillation (\method{}), pools the visual tokens at the input of the language model without executing the decoder, and supervises a closed-form ridge adapter with targets that mix language-derived class anchors with orthogonal label codes. We show that language guidance adds no cost to the closed-form solve, which depends on the language model only through a class-level kernel. On ImageNet-100 with one image per class, \method{} is within 2.41 points of full-data training, compared with 5.6 points for the strongest prior method. Compared with gradient matching on the encoder parameters of the same VLMs, it is more accurate, up to $21.5\times$ faster, and transfers better to unseen encoders.
\end{abstract}

\section{Introduction}
\label{sec:intro}

Large-scale pre-training has changed how visual recognition models are built. A pre-trained backbone is kept frozen, and only a lightweight head is fitted for each downstream task. Dataset distillation (DD)~\cite{wang2018dataset,yu2024dataset} in this regime seeks a small synthetic set on which the fitted head performs as well as a head fitted on the full training set. Linear Gradient Matching (LGM)~\cite{cazenavette2025lgm} introduced this problem and showed that a single synthetic image per class, optimized by matching linear-probe gradients, outperforms real-image selection. CLP-DD~\cite{peng2026clpdd} replaced gradient matching with the closed-form ridge solution of the induced probe, which reduces distillation time by an order of magnitude.

\begin{figure}[t]
    \centering
    \includegraphics[width=\linewidth]{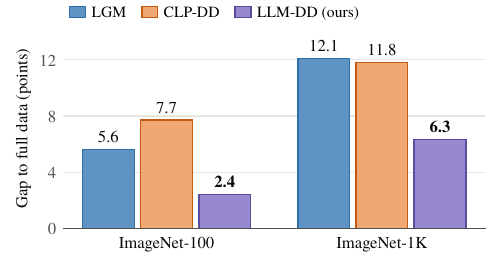}
    \caption{Gap in Top-1 accuracy between one image per class and full-data training (lower is better), averaged over the backbones of each method. LGM~\cite{cazenavette2025lgm} and CLP-DD~\cite{peng2026clpdd} use four ViT-B encoders, and \method{} uses Qwen3-VL and DeepSeek-VL2.}
    \label{fig:teaser}
\end{figure}

However, with one image per class, the strongest reported results remain, on average over four encoders, 5.6 points below the full-data reference on ImageNet-100~\cite{cazenavette2025lgm} and 11.8 points below it on ImageNet-1K~\cite{peng2026clpdd} (Figure~\ref{fig:teaser}). Moreover, both methods supervise the probe with one-hot labels, which treat every pair of classes as equally different, so the distilled set conveys little about how the classes are related to one another. This information is available in large vision--language models (VLMs)~\cite{liu2023visual} such as Qwen3-VL~\cite{bai2025qwen3vl} and DeepSeek-VL2~\cite{wu2024deepseekvl2}, whose language model represents every class name in a space that encodes such relationships, for example that two bird species are closer to each other than either is to a vehicle, and whose visual tokens are aligned with the input of this language model.

Such a VLM, though, cannot be inserted into existing distillation pipelines, which assume an encoder that returns one embedding per image, and it raises three obstacles. The first is the interface: a generative VLM converts an image into a sequence of visual tokens whose length varies with the input resolution and passes them to a large language decoder. No global image embedding is available, and differentiating through the decoder for every synthetic image is prohibitively expensive. The second is supervision: unlike contrastive models~\cite{radford2021clip}, a generative VLM has no text embedding aligned with a pooled visual feature, so class names cannot serve as a classifier directly. When text representations are used as regression targets instead, semantically similar classes receive nearly identical targets and become harder to separate. The third is cost: language targets have the width of the language model, so a closed-form adapter must be solved for thousands of target coordinates instead of one per class.

We propose Language--Label Mixing Dataset Distillation (\method{}), which resolves these obstacles within the closed-form framework. First, a token-pooled interface fixes the input grid, pools the visual tokens at the input of the language model into a single feature, and does not execute the decoder on images. Second, language--label mixing builds targets that interpolate between language-derived class anchors and orthogonal label codes with a single coefficient. The language part carries the relationships between classes, and the orthogonal part bounds the pairwise target distance and the condition number of the target kernel for any text anchors. Third, we prove that closed-form ridge adaptation to these targets, followed by cosine classification, depends on the language representations only through a $C\times C$ class kernel. The solve therefore has $C$ right-hand sides regardless of the width of the language model, and the kernel can accompany the images to adapt encoders that have no text tower. The synthetic images are optimized with a margin-based discriminative objective on real features.

On ImageNet-100 with one image per class, \method{} reaches 92.35\% with Qwen3-VL and 90.89\% with DeepSeek-VL2, 1.98 and 2.84 points below full-data training, whereas the smallest gap reported for a ViT-B encoder is 3.7 points~\cite{cazenavette2025lgm}. On ImageNet-1K, the gaps are 5.84 and 6.84 points, compared with at least 7.4 points in prior work~\cite{peng2026clpdd}. At five images per class, the gap on Qwen3-VL decreases to 0.21 points. On the same VLM backbones, \method{} improves over gradient matching on the encoder parameters by 6.62 points on ImageNet-100 while being up to $21.5\times$ faster (Figure~\ref{fig:efficiency}), and its images transfer better to four unseen ViT-B encoders (50.61\% versus 45.57\%).

Our contributions are summarized as follows:
\begin{itemize}
    \item We extend dataset distillation for pre-trained models from vision encoders to large generative VLMs, and identify three obstacles specific to these models: the absence of a global image embedding, the absence of an aligned text embedding, and the width of language targets.
    \item We propose \method{}, which combines a token-pooled visual interface, language--label mixing with a guaranteed minimum class separation, and an exact class-space reformulation whose cost is independent of the language-model width and whose $C\times C$ kernel is the only language-side quantity required to adapt another encoder.
    \item With Qwen3-VL and DeepSeek-VL2, \method{} reduces the gap to full-data training at one image per class to 2.41 points on ImageNet-100 and 6.34 points on ImageNet-1K, smaller than every gap reported for ViT-B encoders. On the same backbones, it outperforms gradient matching on the encoder parameters in accuracy, in optimization time by up to $21.5\times$, and in transfer to unseen encoders.
\end{itemize}

\section{Related Work}
\label{sec:related}

\paragraph{Dataset Distillation.}
DD methods~\cite{wang2018dataset,yu2024dataset} optimize synthetic data by matching gradients~\cite{zhao2021gradient,zhao2021dsa}, training trajectories~\cite{cazenavette2022trajectory,guo2024towards}, or feature distributions~\cite{zhao2023distribution,wang2022cafe}. Parameter-matching methods can prune parameters that are difficult to match~\cite{li2024pruning} or learn how much weight each parameter receives~\cite{li2024iadd}. Distribution matching has also been extended to hyperbolic features to capture hierarchy~\cite{li2025hyperbolic} and to statistics conditioned on sample difficulty~\cite{ma2026difficulty}. Kernel methods fit predictors on synthetic examples by ridge regression~\cite{nguyen2021kip,nguyen2021wide,loo2022efficient,zhou2022frepo}.

Other approaches change how the distilled images are represented or generated~\cite{kim2022dataset,cazenavette2023glad,gu2024efficient}. Generative methods balance global structure with local detail~\cite{li2024global}, use a memory to track distribution alignment and guide diffusion sampling~\cite{li2025memory}, or select generated examples by classification difficulty~\cite{li2025sampling}. Color compensation combines real-image selection with diffusion-based refinement~\cite{wu2025color}. For fine-grained classes, FD$^2$ uses class prototypes and attention constraints~\cite{ma2026fd2}, while DeCO composes images from discriminative patches of the same class~\cite{fan2026deco}. Several methods have scaled distillation to ImageNet-1K~\cite{cui2023scaling,yin2023squeeze,sun2024diversity}. Most of these methods are designed for training networks from scratch.

\paragraph{Dataset Distillation for Pre-trained Models.}
LGM~\cite{cazenavette2025lgm} formulated DD for frozen pre-trained encoders such as CLIP~\cite{radford2021clip}, DINOv2~\cite{oquab2024dinov2}, MoCo-v3~\cite{chen2021mocov3}, and EVA-02~\cite{fang2024eva02} by matching the gradients of randomly initialized linear probes under repeated differentiable augmentation. CLP-DD~\cite{peng2026clpdd} computes the induced probe in closed form and is about $14\times$ faster at comparable ImageNet-1K accuracy. Both use ViT-B encoders and one-hot labels, which \method{} replaces with large VLMs and language-derived targets.

\paragraph{Vision--Language Models and Language Supervision.}
Contrastive models~\cite{radford2021clip} place images and text in one embedding space, which supports class-name classifiers, lightweight adapters~\cite{gao2024clipadapter,zhang2022tipadapter}, and learned prompts~\cite{zhou2022learning}. Generative VLMs~\cite{liu2023visual,bai2025qwen3vl,wu2024deepseekvl2} instead project visual tokens into the input space of a language model. Earlier work paired distilled medical images with soft labels~\cite{li2022soft}. Language has entered dataset distillation through image--text pair distillation~\cite{wu2024vldd} and through vision--language prototypes or fusion in generative pipelines~\cite{zou2025vlcp,cai2026evlf}. In \method{}, the images remain the only learned variables, and language shapes the adapter targets, as in semantic label embeddings~\cite{frome2013devise}.

\section{Method}
\label{sec:method}

We consider dataset distillation with a frozen generative VLM that consists of a vision tower, a connector that maps patch features into the input space of the language model, and the language model itself. All parameters are fixed throughout distillation. Given real training data $\mathcal{D}$ with $C$ classes, we optimize a synthetic set $\mathcal{S}$ of $n=C\times\mathrm{IPC}$ images whose fixed labels are encoded by the one-hot matrix $H\in\{0,1\}^{n\times C}$. Our goal is to optimize $\mathcal{S}$ such that an adapter fitted on its features performs well on real data.

\begin{figure*}[t]
    \centering
    \includegraphics[width=\linewidth]{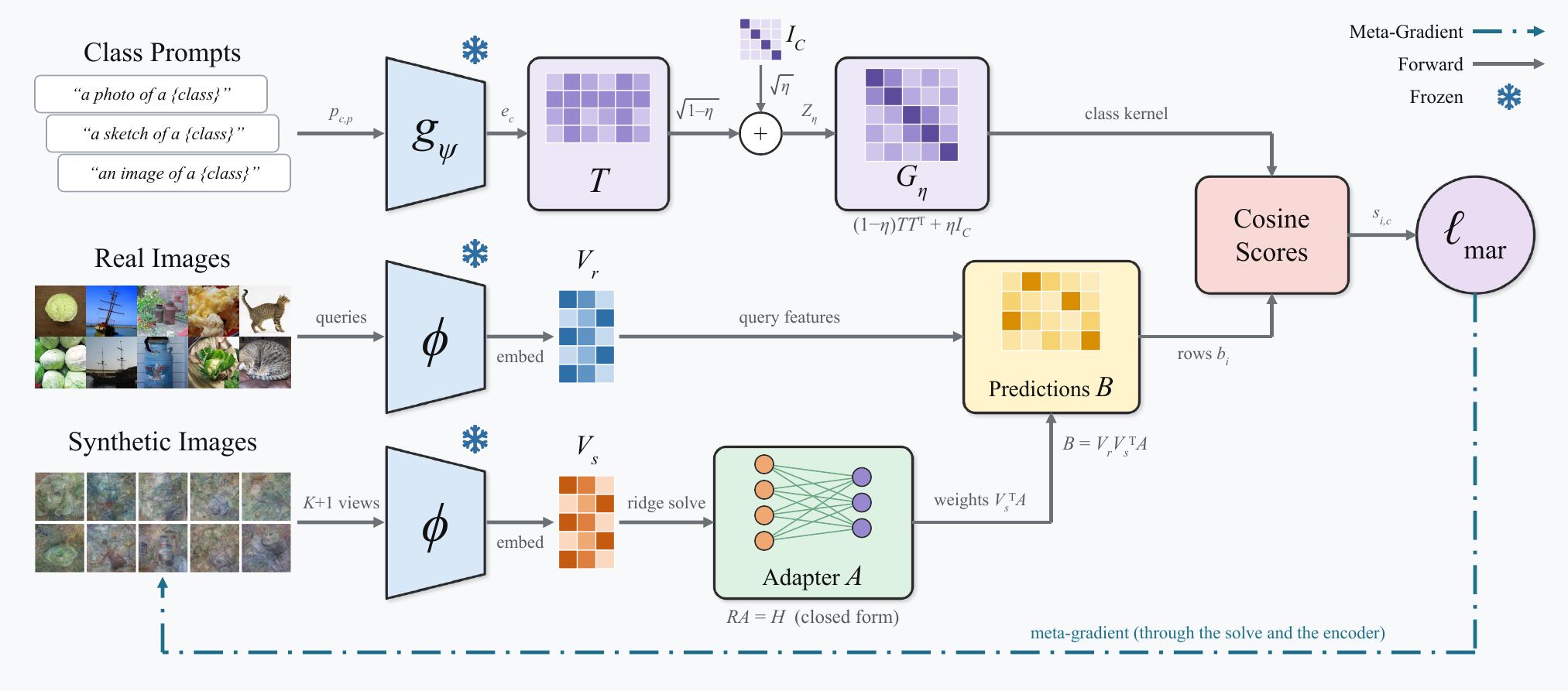}
    \caption{Overview of \method{}. Class prompts are encoded once by the language model $g_\psi$ into anchors $T$, which are concatenated ($\oplus$) with the scaled label codes $I_C$ into the targets $Z_\eta$ and the class kernel $G_\eta=Z_\eta Z_\eta^\top$. Real and synthetic images are embedded by the frozen visual interface $\phi$. The ridge adapter $A$ is solved in closed form on the synthetic features $V_s$ and applied to the real features $V_r$, and the predictions $B$ are scored against the classes through $G_\eta$. The margin loss $\ell_{\mathrm{mar}}$ is backpropagated through the solve and the encoder to the synthetic images.}
    \label{fig:method}
\end{figure*}

\method{} has four components (Figure~\ref{fig:method}): a token-pooled interface that provides differentiable visual features and class anchors (Section~\ref{sec:interface}), language--label mixing that converts the anchors into well-separated targets (Section~\ref{sec:targets}), a closed-form adapter solved in class space (Section~\ref{sec:adapter}), and a margin-based objective on real features that updates the synthetic images (Section~\ref{sec:objective}). The full procedure is given in Algorithm~\ref{alg:llmdd}.

\subsection{Token-Pooled VLM Interface}
\label{sec:interface}

\paragraph{Visual features.}
Encoders used in prior work return one embedding per image. A generative VLM returns a sequence of visual tokens whose length depends on the image resolution, through native dynamic resolution in Qwen3-VL~\cite{bai2025qwen3vl} and dynamic tiling in DeepSeek-VL2~\cite{wu2024deepseekvl2}, and consumes them with its language decoder. We fix the input to a single grid of constant resolution, so that every image yields the same number $M$ of tokens $u_1(x),\dots,u_M(x)\in\mathbb{R}^{d_v}$ at the output of the connector, and define
\begin{equation}
    \phi(x)=\mathcal{N}\Big(M^{-1}\sum_{m=1}^{M}u_m(x)\Big),
    \label{eq:visual_feature}
\end{equation}
where $\mathcal{N}(v)=v/\max(\|v\|_2,\epsilon)$ and $\epsilon>0$ is a numerical safeguard. This interface has three properties. The feature has a fixed size and is differentiable with respect to the pixels, as required for optimizing synthetic images. The tokens are taken where the language model reads them, in a space that pre-training has aligned with language. The language decoder is not executed on images, so the forward and backward passes of distillation involve only the vision tower and the connector. We denote by $V_s=\phi(\mathcal{S})\in\mathbb{R}^{n\times d_v}$ the features of the synthetic images and by $V_r\in\mathbb{R}^{N_q\times d_v}$ those of a batch of $N_q$ real queries.

\paragraph{Language anchors.}
The language model of a generative VLM has no embedding head for sentences, and the hidden states of different prompts share a dominant common component~\cite{ethayarajh2019contextual}, which makes raw class representations nearly parallel. For each of the $P=4$ prompts of class $c$, the representation $g_\psi(p_{c,p})$ is the mean of the final-layer hidden states of the language model over all non-padding tokens. We average the prompt representations, remove the cross-class mean, and normalize:
\begin{equation}
    \begin{aligned}
        e_c &= P^{-1}\sum_{p=1}^{P} g_\psi(p_{c,p}), \\
        t_c &= \mathcal{N}\Big(e_c - C^{-1}\sum_{j=1}^{C} e_j\Big).
    \end{aligned}
    \label{eq:class_anchors}
\end{equation}
Centering removes the shared component and leaves the directions along which classes differ. The rows of $T\in\mathbb{R}^{C\times d_t}$ are these unit-norm anchors, where $d_t$ is the width of the language model. The language model is executed once on $CP$ prompts and is not needed afterwards.

\subsection{Language--Label Mixing}
\label{sec:targets}

The anchors cannot replace a classifier, since they are not aligned with $\phi$, and they are imperfect regression targets, since anchors of semantically similar classes are close and the adapter must map different classes to nearly the same point. We therefore mix them with orthogonal label codes:
\begin{equation}
    \begin{aligned}
        Z_\eta &= \big[\sqrt{1-\eta}\,T,\ \sqrt{\eta}\,I_C\big], \\
        G_\eta &= Z_\eta Z_\eta^\top = (1-\eta)\,TT^\top + \eta I_C,
    \end{aligned}
    \label{eq:mixed_targets}
\end{equation}
where $0\le\eta\le 1$ and $d_z=d_t+C$. We omit the subscript $\eta$ when unambiguous. The coefficient $\eta$ controls how much the targets rely on language: $\eta=0$ uses the anchors alone, and $\eta=1$ reduces to the one-hot labels of prior work. Each mixed target $z_c$ has unit norm, and, for $i\neq j$,
\begin{equation}
    \|z_i-z_j\|_2^2 = (1-\eta)\|t_i-t_j\|_2^2 + 2\eta ,
    \label{eq:target_distance}
\end{equation}
so any $\eta>0$ keeps two classes apart even when their anchors coincide. The class kernel $G$ is positive definite for $\eta>0$, with an explicit bound on its condition number. If $\mu_j$ are the eigenvalues of $TT^\top$, the eigenvalues of $G$ are $\eta+(1-\eta)\mu_j$, and since $\operatorname{tr}(TT^\top)=C$,
\begin{equation}
    \lambda_{\min}(G)\ge\eta, \qquad
    \kappa_2(G)\le 1+\frac{C(1-\eta)}{\eta} .
    \label{eq:kernel_bounds}
\end{equation}
A small $\eta$ thus preserves the semantic relationships between classes while guaranteeing a minimum separation.

\subsection{Closed-Form Adapter in Class Space}
\label{sec:adapter}

The inner problem fits a linear adapter $W\in\mathbb{R}^{d_v\times d_z}$ that maps synthetic features to the targets of their classes with the ridge objective~\cite{hoerl1970ridge}
\begin{equation}
    W^\ast = \arg\min_W\ \frac{1}{2}\|V_sW-HZ\|_F^2
    + \frac{\lambda}{2}\|W\|_F^2, \qquad \lambda>0 .
    \label{eq:ridge_objective}
\end{equation}
The problem is strongly convex, and its unique solution in sample-space form is
\begin{equation}
    W^\ast=V_s^\top R^{-1}HZ, \qquad R=V_sV_s^\top+\lambda I_n .
    \label{eq:adapter_solution}
\end{equation}
Evaluating Eq.~\eqref{eq:adapter_solution} as written requires solving for all $d_z=d_t+C$ target coordinates, a number dominated by the width of the language model. The following result shows that this is unnecessary for cosine classification.

\begin{llmddprop}[Exact class-space equivalence]
\label{prop:class_space}
For fixed targets in Eq.~\eqref{eq:mixed_targets}, define
\begin{equation}
    RA=H, \qquad B=V_rV_s^\top A .
    \label{eq:class_space_solve}
\end{equation}
The predictions of the adapter on the queries are $BZ$. For row $b_i$ of $B$, their cosine scores against the targets are
\begin{equation}
    s_i=\frac{b_iG}{\max\!\big(\sqrt{b_iGb_i^\top},\,\epsilon\big)} .
    \label{eq:class_space_scores}
\end{equation}
Consequently, any loss on these scores and the labels, and its gradient with respect to the synthetic features, depend on the targets only through $G$.
\end{llmddprop}

Appendix~\ref{app:proof} gives the proof. Proposition~\ref{prop:class_space} has three consequences. First, once $R$ is factorized, the solve in Eq.~\eqref{eq:class_space_solve} has $C$ right-hand sides instead of $d_z$, which reduces its cost from $O(n^2d_z)$ to $O(n^2C)$. The cost is the same as with one-hot labels and does not grow with the width of the language model. Second, the language model enters distillation only through the $C\times C$ kernel $G$, which is computed once. Third, the same holds at adaptation time. The distilled dataset consists of the images $\mathcal{S}$ and the kernel $G$, and any frozen encoder $\phi'$ can be adapted by solving Eq.~\eqref{eq:class_space_solve} on $\phi'(\mathcal{S})$ and scoring with Eq.~\eqref{eq:class_space_scores}. The encoder $\phi'$ needs no text tower, which allows the class semantics of the source VLM to be used with vision-only models.

\subsection{Margin-Based Outer Objective}
\label{sec:objective}

To optimize the images, we evaluate the adapter induced by the current synthetic images on real training features. Given the scores $s_{i,c}$ of query $i$ with label $y_i$, we apply an additive cosine margin~\cite{wang2018cosface,deng2019arcface} to the true class:
\begin{equation}
    \begin{aligned}
        \ell_{i,c} &= \frac{s_{i,c}-m\,\mathbf{1}[c=y_i]}{\tau}, \\
        \mathcal{L}_{\mathrm{mar}} &= -\frac{1}{N_q}\sum_{i}
        \log\frac{e^{\ell_{i,y_i}}}{\sum_{c} e^{\ell_{i,c}}} ,
    \end{aligned}
    \label{eq:margin_loss}
\end{equation}
where $m\ge 0$ is the margin and $\tau>0$ the temperature. The loss encourages each real sample to score higher on the target of its own class than on every competing class by a margin $m$, and the margin is used during distillation only. Compared with a squared loss on the targets, this objective directly optimizes the class discrimination that determines the accuracy of the adapter. We additionally use an alignment term $\mathcal{L}_{\mathrm{aux}}=N_q^{-1}\sum_i(1-s_{i,y_i})$ with weight $\beta$.

To obtain images that remain informative under image transformations, we evaluate the objective on a clean branch $k=0$ and $K$ augmented branches. Each branch forms its own system $R_kA_k=H$, and the outer objective is
\begin{equation}
    \mathcal{L}_{\mathrm{out}} = \sum_{k=0}^{K}\omega_k
    \Big(\mathcal{L}_{\mathrm{mar}}^{(k)}
    +\beta\mathcal{L}_{\mathrm{aux}}^{(k)}\Big)
    +\gamma\,\mathrm{TV}(\mathcal{S}),
    \label{eq:outer_objective}
\end{equation}
where $\omega_k\ge 0$, $\sum_k\omega_k=1$, and $\mathrm{TV}(\mathcal{S})$ is a total-variation regularizer~\cite{rudin1992nonlinear,mahendran2015understanding} with weight $\gamma\ge 0$. The gradient of $\mathcal{L}_{\mathrm{out}}$ with respect to the synthetic features has a closed form that requires one additional solve with the same matrix $R$ (see Appendix~\ref{app:adjoint}). It is propagated through the vision tower to the images, which use a color-decorrelated pyramid parameterization~\cite{cazenavette2025lgm,olah2017feature}.

\section{Experiments}
\label{sec:experiments}

We evaluate \method{} on ImageNet-100~\cite{tian2020contrastive} and ImageNet-1K~\cite{russakovsky2015imagenet}. The dataset variants and the ImageNet-100 class subset follow CLP-DD~\cite{peng2026clpdd}. The source models are Qwen3-VL-4B-Instruct~\cite{bai2025qwen3vl} and DeepSeek-VL2-Tiny~\cite{wu2024deepseekvl2}, referred to as \qwenname{} and \deepseekname{}. All pre-trained parameters remain frozen. Unless otherwise stated, the budget is one image per class (IPC${}=1$), and experiments are run on a single NVIDIA RTX A6000 GPU (48\,GB), or an NVIDIA H100 GPU for ImageNet-1K. Further settings are given in Appendix~\ref{app:experiment_settings}.

\paragraph{Evaluation protocol.}
The metric is the Top-1 accuracy of the adapter induced by the final distilled images. For every method run on the VLM backbones, we fit the closed-form adapter of Section~\ref{sec:adapter} on the frozen features of its images, with the same class kernel and ridge coefficient and with margin-free cosine scoring, and evaluate it on the held-out split. We report the mean and standard deviation over three independently distilled sets. Hyperparameters are selected on a development split derived from the training set, with a fixed distillation endpoint and no checkpoint selection on the evaluation split. Results of LGM~\cite{cazenavette2025lgm} and CLP-DD~\cite{peng2026clpdd} on ViT-B encoders are those reported in the respective papers with a trained linear probe, and are compared through the gap to the full-data reference of each backbone. For a comparison on the same backbone, we also include gradient-matching and feature-matching baselines in the same pipeline.

\subsection{Comparison with Prior Work}
\label{sec:prior_work}

\begin{table}[t]
    \centering
    \small
    \setlength{\tabcolsep}{2.8pt}
    \caption{Comparison with distillation methods for pre-trained models at IPC${}=1$. Top-1 accuracy (\%) with one image per class, with the full dataset, and their difference (lower is better). Results of LGM and CLP-DD are taken from~\cite{cazenavette2025lgm} and~\cite{peng2026clpdd}.}
    \label{tab:prior_work}
    \begin{tabular}{clcccccc}
        \toprule
        & & \multicolumn{3}{c}{ImageNet-100} & \multicolumn{3}{c}{ImageNet-1K} \\
        \cmidrule(lr){3-5}
        \cmidrule(lr){6-8}
        & Backbone & IPC${}=1$ & Full & Gap\,$\downarrow$ & IPC${}=1$ & Full & Gap\,$\downarrow$ \\
        \midrule
        \multirow{5}{*}{\rotatebox[origin=c]{90}{LGM}}
        & CLIP      & 84.9 & 92.5 & 7.6  & 63.0 & 78.7 & 15.7 \\
        & DINOv2    & 91.5 & 95.2 & 3.7  & 75.0 & 83.0 & 8.0 \\
        & EVA-02    & 89.0 & 94.1 & 5.1  & 70.3 & 81.7 & 11.4 \\
        & MoCo-v3   & 83.4 & 89.4 & 6.0  & 63.2 & 76.5 & 13.3 \\
        & Average   & 87.2 & 92.8 & 5.6  & 67.9 & 80.0 & 12.1 \\
        \midrule
        \multirow{5}{*}{\rotatebox[origin=c]{90}{CLP-DD}}
        & CLIP      & 79.2 & 92.4 & 13.2 & 62.4 & 78.7 & 16.3 \\
        & DINOv2    & 89.2 & 95.1 & 5.9  & 75.6 & 83.0 & 7.4 \\
        & EVA-02    & 87.6 & 94.0 & 6.4  & 71.2 & 81.7 & 10.5 \\
        & MoCo-v3   & 83.9 & 89.3 & 5.4  & 63.4 & 76.5 & 13.1 \\
        & Average   & 85.0 & 92.7 & 7.7  & 68.2 & 80.0 & 11.8 \\
        \midrule
        \multirow{3}{*}{\rotatebox[origin=c]{90}{Ours}}
        & \qwenname{}     & \textbf{92.35} & 94.33 & \textbf{1.98} & \textbf{76.28} & 82.12 & \textbf{5.84} \\
        & \deepseekname{} & 90.89 & 93.73 & 2.84 & 73.33 & 80.17 & 6.84 \\
        & Average         & 91.62 & 94.03 & 2.41 & 74.80 & 81.14 & 6.34 \\
        \bottomrule
    \end{tabular}
\end{table}

Table~\ref{tab:prior_work} compares \method{} with the two existing methods for pre-trained models. On ImageNet-100, \method{} reaches 92.35\% with \qwenname{}, above the highest accuracy reported at one image per class (91.5\% for LGM with DINOv2). The gap to full-data training is 1.98 points on \qwenname{} and 2.84 points on \deepseekname{}, whereas it ranges from 3.7 to 7.6 points for LGM and from 5.4 to 13.2 points for CLP-DD. On ImageNet-1K, \method{} reaches 76.28\% with \qwenname{}, compared with 75.0\% for LGM and 75.6\% for CLP-DD with DINOv2, and its gaps of 5.84 and 6.84 points are below the smallest gap of either method (7.4 points). Averaged over backbones, \method{} reduces the gap of the strongest prior result from 5.6 to 2.41 points on ImageNet-100 and from 11.8 to 6.34 points on ImageNet-1K. The full-data reference of the VLMs is on average about one point higher than that of the ViT-B encoders, so the smaller gap is obtained against a stronger reference. Sections~\ref{sec:standard_evaluation} and~\ref{sec:ablation} then examine the method with the backbone and the evaluation head held fixed.

\subsection{Controlled Comparison on VLM Backbones}
\label{sec:standard_evaluation}

\begin{figure*}[t]
    \centering
    \includegraphics[width=0.8\linewidth]{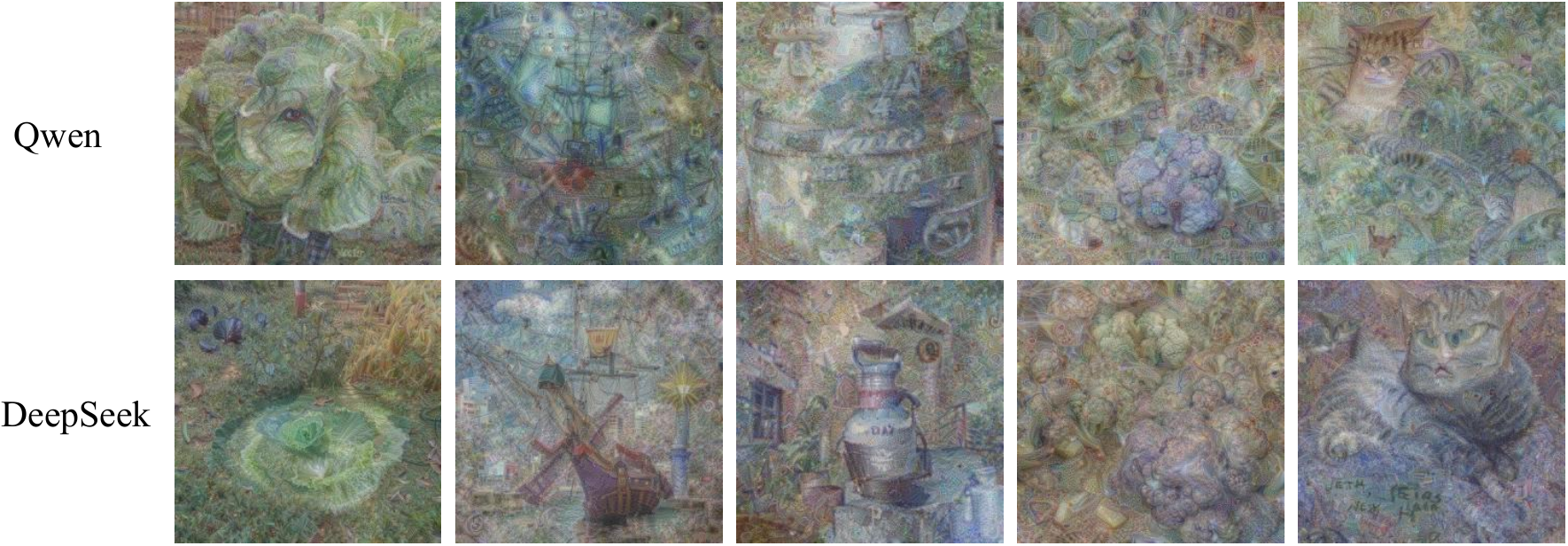}
    \caption{Examples of IPC${}=1$ images distilled by \method{} on ImageNet-100. Rows correspond to the \qwenname{} and \deepseekname{} backbones. Columns show, from left to right, head cabbage, pirate, milk can, cauliflower, and tabby.}
    \label{fig:distilled_in100}
\end{figure*}

\begin{table*}[t]
    \centering
    \small
    \setlength{\tabcolsep}{6pt}
    \caption{Controlled comparison at IPC${}=1$ with the same backbone, pipeline, and evaluation head. GM is class-wise gradient matching with respect to the visual and head parameters and FM is feature matching, both in the pipeline of \method{}. Entries are the mean $\pm$ standard deviation of Top-1 accuracy (\%) over three runs, and Avg.\ averages the two backbones.}
    \label{tab:imagenet_main}
    \begin{tabular}{lccc>{}c>{}c>{}c}
        \toprule
        & \multicolumn{3}{c}{ImageNet-100}
        & \multicolumn{3}{c}{ImageNet-1K} \\
        \cmidrule(lr){2-4}
        \cmidrule(lr){5-7}
        Method
        & \qwenname{} & \deepseekname{} & Avg.
        & \qwenname{} & \deepseekname{} & Avg. \\
        \midrule
        Random
        & 68.83\pmstd{2.5} & 64.96\pmstd{1.2} & 66.90\pmstd{1.8}
        & 38.79\pmstd{1.8} & 36.83\pmstd{2.2} & 37.81\pmstd{2.1} \\
        Centroid
        & 78.25\pmstd{0.1} & 76.19\pmstd{0.2} & 77.22\pmstd{0.2}
        & 47.23\pmstd{0.5} & 43.76\pmstd{0.3} & 45.49\pmstd{0.4} \\
        Neighbor
        & 66.36\pmstd{0.5} & 59.45\pmstd{0.7} & 62.91\pmstd{0.6}
        & 40.13\pmstd{0.4} & 35.97\pmstd{0.9} & 38.05\pmstd{0.6} \\
        \midrule
        GM
        & 88.23\pmstd{0.8} & 81.77\pmstd{1.5} & 85.00\pmstd{1.1}
        & 69.37\pmstd{1.2} & 67.42\pmstd{1.9} & 68.39\pmstd{1.6} \\
        FM
        & 85.97\pmstd{0.3} & 82.94\pmstd{0.7} & 84.45\pmstd{0.5}
        & 66.73\pmstd{0.2} & 64.32\pmstd{0.5} & 65.52\pmstd{0.4} \\
        \method{}
        & \textbf{92.35}\pmstd{0.1} & \textbf{90.89}\pmstd{0.3}
        & \textbf{91.62}\pmstd{0.2}
        & \textbf{76.28}\pmstd{0.1} & \textbf{73.33}\pmstd{0.3} & \textbf{74.80}\pmstd{0.2} \\
        \midrule
        Full data
        & 94.33\pmstd{0.1} & 93.73\pmstd{0.0} & 94.03\pmstd{0.0}
        & 82.12\pmstd{0.1} & 80.17\pmstd{0.0} & 81.14\pmstd{0.0} \\
        \bottomrule
    \end{tabular}
\end{table*}

Table~\ref{tab:imagenet_main} compares methods on the same VLM backbone: Random samples real training images within each class, Centroid selects the images nearest to the class mean feature, and Neighbor retrieves the real images nearest to the \method{} images. GM applies gradient matching~\cite{zhao2021gradient,zhao2021dsa} to the frozen VLM: for each class, the gradients of a cross-entropy loss with respect to the visual encoder, the connector, and a randomly initialized linear head are matched between synthetic and real images, without updating any weights. FM matches each synthetic feature to the class-wise mean of real features~\cite{zhao2023distribution}. Both replace only the objective of \method{} and keep its feature interface, image parameterization, initialization, augmentation branches, real-data sampling, and update budget. With the same augmentation, matching the linear-head gradients alone, as in LGM~\cite{cazenavette2025lgm}, gives lower mean accuracy than GM across the two VLMs on ImageNet-100 (Table~\ref{tab:app_lgm}).

On ImageNet-100, \method{} improves the two-backbone average over GM by 6.62 points (91.62\% versus 85.00\%), FM by 7.17 points, and Centroid by 14.40 points. On ImageNet-1K, the corresponding gains are 6.41, 9.28, and 29.31 points (74.80\% versus 68.39\%, 65.52\%, and 45.49\%). GM is the stronger baseline on \qwenname{} and FM on \deepseekname{} for ImageNet-100, whereas \method{} exceeds both on each backbone and dataset. Since GM and FM share the pipeline of \method{}, these gains are attributable to the objective. Figure~\ref{fig:distilled_in100} shows recognizable class-relevant content for both backbones: \qwenname{} produces dense, class-related patterns, while \deepseekname{} tends to place individual objects in scene-like contexts. Appendix~\ref{app:additional} analyzes the feature space, and Appendix~\ref{app:additional_distilled_images} shows further images.

\subsection{Cross-Model Transfer}
\label{sec:cross_model}

\begin{table}[t]
    \centering
    \small
    \setlength{\tabcolsep}{4.5pt}
    \caption{Cross-model transfer on ImageNet-100 at IPC${}=1$. Images are distilled with \qwenname{} and evaluated on a different frozen encoder. GM and FM are the baselines of Table~\ref{tab:imagenet_main}, and real images are random training images at the same budget.}
    \label{tab:cross_model}
    \begin{tabular}{lcccc}
        \toprule
        Method & CLIP & DINOv2 & EVA-02 & MoCo-v3 \\
        \midrule
        GM
        & 43.75\pmstd{0.1} & 51.43\pmstd{0.3} & 49.32\pmstd{0.1} & 37.79\pmstd{0.3} \\
        FM
        & 32.17\pmstd{0.2} & 43.33\pmstd{0.1} & 37.19\pmstd{0.6} & 31.92\pmstd{0.1} \\
        \method{}
        & \textbf{46.92}\pmstd{0.2} & \textbf{59.32}\pmstd{0.2} & \textbf{54.14}\pmstd{0.3} & \textbf{42.04}\pmstd{0.1} \\
        \midrule
        Real images
        & 54.43\pmstd{0.1} & 77.82\pmstd{0.3} & 68.77\pmstd{0.1} & 59.96\pmstd{0.1} \\
        \bottomrule
    \end{tabular}
\end{table}

We evaluate whether images distilled with a VLM can be used to adapt other models. The ImageNet-100 sets distilled with \qwenname{} at IPC${}=1$ are evaluated without modification on the four ViT-B encoders of prior work, CLIP~\cite{radford2021clip}, DINOv2~\cite{oquab2024dinov2}, EVA-02~\cite{fang2024eva02}, and MoCo-v3~\cite{chen2021mocov3}, none of which is used during distillation. Each target encoder is adapted with the closed-form head on its own features of the images, using the class kernel $G$ of the source model (Section~\ref{sec:adapter}), so the vision-only targets require no text tower.

Table~\ref{tab:cross_model} reports the results. \method{} transfers better than GM on every target encoder, with an average of 50.61\% compared with 45.57\%, and the gain is largest on DINOv2 (7.89 points) and EVA-02 (4.82 points). FM transfers least (36.15\%), 14.46 points below \method{}. The 50.61\% average of \method{}, obtained across model families, is comparable to the 50.9\% to 52.3\% reported for CLP-DD between ViT-B encoders~\cite{peng2026clpdd}. However, real images at the same budget remain more transferable (65.2\% on average), so the distilled images remain specific to the source model.

\subsection{IPC Scaling}
\label{sec:ipc_scaling}

\begin{table}[t]
    \centering
    \small
    \setlength{\tabcolsep}{4.5pt}
    \caption{IPC scaling on ImageNet-100. Columns give the IPC. Each backbone is used for both distillation and evaluation.}
    \label{tab:ipc_scaling}
    \begin{tabular}{lcccccc}
        \toprule
        & \multicolumn{3}{c}{\qwenname{}}
        & \multicolumn{3}{c}{\deepseekname{}} \\
        \cmidrule(lr){2-4}
        \cmidrule(lr){5-7}
        Method & 1 & 3 & 5 & 1 & 3 & 5 \\
        \midrule
        Random      & 68.83 & 84.14 & 88.29 & 64.96 & 80.44 & 85.51 \\
        Centroid    & 78.25 & 86.35 & 88.41 & 76.19 & 83.22 & 86.27 \\
        FM & 85.97 & 89.22 & 90.43 & 82.94 & 86.17 & 88.79 \\
        \method{}
        & \textbf{92.35} & \textbf{93.79} & \textbf{94.12}
        & \textbf{90.89} & \textbf{92.19} & \textbf{92.94} \\
        \bottomrule
    \end{tabular}
\end{table}

Table~\ref{tab:ipc_scaling} reports ImageNet-100 results at IPC $\in\{1,3,5\}$, where every budget is optimized independently. \method{} outperforms Random, Centroid, and FM at each IPC on both backbones, and \method{} at IPC${}=1$ exceeds every baseline of Table~\ref{tab:ipc_scaling} at IPC${}=5$. The gains over FM are 6.38, 4.57, and 3.69 points on \qwenname{} and 7.95, 6.02, and 4.15 points on \deepseekname{} at IPC${}=1$, 3, and 5. At IPC${}=5$, \method{} is within 0.21 points of the full-data reference on \qwenname{} (94.12\% versus 94.33\%) and within 0.79 points on \deepseekname{} (92.94\% versus 93.73\%). For comparison, CLP-DD with DINOv2 reaches 91.8\% at IPC${}=5$, 3.3 points below its full-data reference~\cite{peng2026clpdd}. From IPC${}=1$ to 5, Random improves by about 20 points on each backbone, whereas \method{} improves by 1.77 points on \qwenname{} and 2.05 points on \deepseekname{}, so most of its accuracy is already reached with one image per class.

\subsection{Efficiency Analysis}
\label{sec:efficiency}

\begin{figure*}[t]
    \centering
    \includegraphics[width=\linewidth]{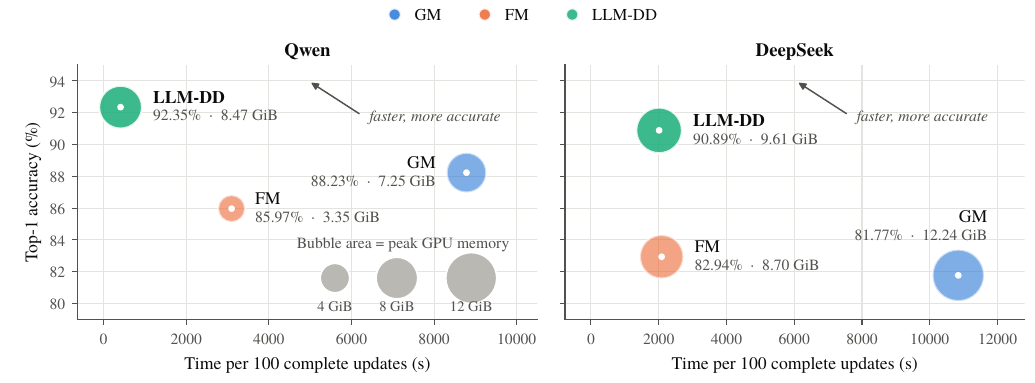}
    \caption{Accuracy versus efficiency on ImageNet-100 at IPC${}=1$ on a single NVIDIA RTX A6000 (48\,GB). \method{} attains the highest accuracy at the lowest time cost on both backbones. Bubble area denotes peak GPU memory.}
    \label{fig:efficiency}
\end{figure*}

\begin{table}[t]
    \centering
    \small
    \setlength{\tabcolsep}{3pt}
    \caption{Efficiency on ImageNet-100 at IPC${}=1$: time in seconds per 100 updates, peak memory in GiB, and Top-1 accuracy (\%).}
    \label{tab:efficiency}
    \begin{tabular}{lcccccc}
        \toprule
        & \multicolumn{3}{c}{\qwenname{}}
        & \multicolumn{3}{c}{\deepseekname{}} \\
        \cmidrule(lr){2-4}
        \cmidrule(lr){5-7}
        Method & Time & Mem. & Acc. & Time & Mem. & Acc. \\
        \midrule
        GM
        & 8784.50 & 7.25 & 88.23
        & 10841.19 & 12.24 & 81.77 \\
        FM
        & 3097.41 & \textbf{3.35} & 85.97
        & 2086.80 & \textbf{8.70} & 82.94 \\
        \method{}
        & \textbf{409.37} & 8.47 & \textbf{92.35}
        & \textbf{2011.32} & 9.61 & \textbf{90.89} \\
        \bottomrule
    \end{tabular}
\end{table}

Figure~\ref{fig:efficiency} and Table~\ref{tab:efficiency} compare \method{} with the shared-pipeline baselines in terms of time per 100 complete image updates after warm-up, peak allocated GPU memory, and accuracy, with matched precision, resolution, and real-data sampling. Timing covers feature computation, the objective, backpropagation, and the image update. \method{} is both the fastest and the most accurate method on each backbone. On \qwenname{}, it requires 409.37\,s per 100 updates, a $21.5\times$ speedup over GM, which matches the gradients of the encoder parameters class by class and requires second-order backpropagation through the encoder, and a $7.6\times$ speedup over FM, while improving accuracy by 4.12 and 6.38 points. On \deepseekname{}, \method{} is $5.4\times$ faster than GM and uses 21.5\% less peak memory (9.61 versus 12.24\,GiB), while improving accuracy by 9.12 points. Compared with FM on \deepseekname{}, it has a similar cost and gains 7.95 points. \method{} remains below 10\,GiB on both backbones and uses more peak memory than FM, since the closed-form solve is differentiated for each branch.

\subsection{Background Shift}
\label{sec:background_shift}

\begin{table}[t]
    \centering
    \small
    \setlength{\tabcolsep}{3pt}
    \caption{Background-shift evaluation at IPC${}=1$, following the protocol of LGM~\cite{cazenavette2025lgm}. Entries are means over three runs, and Avg.\ averages the two backbones.}
    \label{tab:background}
    \begin{tabular}{lcccccc}
        \toprule
        & \multicolumn{3}{c}{Spawrious} & \multicolumn{3}{c}{Waterbirds} \\
        \cmidrule(lr){2-4}
        \cmidrule(lr){5-7}
        Method & \qwenname{} & \deepseekname{} & Avg. & \qwenname{} & \deepseekname{} & Avg. \\
        \midrule
        Random & 27.73 & 19.87 & 23.80 & 65.62 & 73.18 & 69.40 \\
        Centroid & 25.37 & 32.07 & 28.72 & 65.03 & 60.99 & 63.01 \\
        Neighbor & 18.55 & 37.41 & 27.98 & 56.08 & 68.00 & 62.04 \\
        GM & 32.33 & 42.17 & 37.25 & 67.12 & 66.21 & 66.67 \\
        FM & 28.78 & 29.62 & 29.20 & 65.08 & 64.72 & 64.90 \\
        \method{} & \textbf{39.41} & \textbf{54.38} & \textbf{46.90} & \textbf{77.84} & \textbf{74.23} & \textbf{76.03} \\
        \midrule
        Full data & 71.46 & 68.00 & 69.73 & 92.80 & 91.27 & 92.03 \\
        \bottomrule
    \end{tabular}
\end{table}

Table~\ref{tab:background} evaluates the distilled sets on Spawrious~\cite{lynch2023spawrious} and Waterbirds~\cite{sagawa2020distributionally}, in which the background--class association differs between training and evaluation. \method{} is the most accurate method on both benchmarks and both backbones. On Spawrious, it reaches 46.90\% on average, compared with 37.25\% for GM and 29.20\% for FM. On Waterbirds, it reaches 76.03\%, compared with 69.40\% for Random, the strongest baseline, and 66.67\% for GM. The margin is smallest on Waterbirds with \deepseekname{} (74.23\% versus 73.18\% for Random). The gaps to full-data training (22.83 and 16.00 points) are larger than on ImageNet.

\subsection{Ablation Study}
\label{sec:ablation}

Ablations are conducted on ImageNet-100 at IPC${}=1$. Each variant changes one factor from the reference configuration and re-optimizes the images with paired initialization seeds, real minibatches, and augmentation draws.

\begin{table}[t]
    \centering
    \small
    \setlength{\tabcolsep}{7pt}
    \caption{Target ablation on ImageNet-100 at IPC${}=1$, from pure language targets ($\eta=0$) to one-hot targets ($\eta=1$).}
    \label{tab:target_geometry}
    \begin{tabular}{lcc}
        \toprule
        Targets & \qwenname{} & \deepseekname{} \\
        \midrule
        $\eta=0$ (language)  & 91.04\pmstd{0.2} & 90.22\pmstd{0.1} \\
        $\eta=0.10$           & 91.78\pmstd{0.3} & 90.46\pmstd{0.0} \\
        $\eta=0.25$
        & \textbf{92.35}\pmstd{0.1} & \textbf{90.89}\pmstd{0.3} \\
        $\eta=0.50$           & 91.54\pmstd{0.1} & 90.26\pmstd{0.1} \\
        $\eta=0.75$           & 91.12\pmstd{0.2} & 89.43\pmstd{0.3} \\
        $\eta=1$ (one-hot) & 89.79\pmstd{0.1} & 88.34\pmstd{0.2} \\
        \bottomrule
    \end{tabular}
\end{table}
\paragraph{Language--label mixing.}
Table~\ref{tab:target_geometry} varies the mixing coefficient $\eta$ in Eq.~\eqref{eq:mixed_targets} between pure language targets ($\eta=0$) and the one-hot targets of prior work ($\eta=1$). The reference value $\eta=0.25$ gives the highest accuracy on both backbones, and accuracy decreases monotonically on either side of it. Compared with one-hot targets, the mixed targets improve accuracy by 2.56 points on \qwenname{} and 2.55 points on \deepseekname{}. Compared with pure language targets, they improve it by 1.31 and 0.67 points. Pure language targets also outperform one-hot targets on both backbones (91.04\% versus 89.79\%, and 90.22\% versus 88.34\%). Language-derived class relationships are therefore useful targets, and a moderate orthogonal component improves them further. With one-hot targets, \method{} remains above the shared-pipeline baselines of Table~\ref{tab:imagenet_main} (89.79\% versus 88.23\% for GM on \qwenname{}, and 88.34\% versus 82.94\% for FM on \deepseekname{}).

\begin{table}[t]
    \centering
    \small
    \setlength{\tabcolsep}{7pt}
    \caption{Component ablations on ImageNet-100 at IPC${}=1$. Each row changes one factor from the reference configuration.}
    \label{tab:component_ablation}
    \begin{tabular}{lcc}
        \toprule
        Variant & \qwenname{} & \deepseekname{} \\
        \midrule
        \method{} (reference)
        & \textbf{92.35}\pmstd{0.1} & \textbf{90.89}\pmstd{0.3} \\
        Outer MSE                  & 90.13\pmstd{0.2} & 88.54\pmstd{0.1} \\
        No margin ($m=0$)          & 92.03\pmstd{0.1} & 90.51\pmstd{0.1} \\
        No alignment ($\beta=0$)   & 91.62\pmstd{0.3} & 89.82\pmstd{0.2} \\
        Clean branch only ($K=0$)  & 90.97\pmstd{0.2} & 89.88\pmstd{0.2} \\
        Single prompt ($P=1$)      & 91.78\pmstd{0.3} & 90.00\pmstd{0.2} \\
        \bottomrule
    \end{tabular}
\end{table}

\paragraph{Outer objective and components.}
Table~\ref{tab:component_ablation} removes one component of \method{} at a time. Outer MSE replaces the classification and alignment terms with a squared target-regression loss while retaining the inner solve and the image regularizer. The remaining variants set the margin to $m=0$, remove the alignment term ($\beta=0$), use only the clean branch ($K=0$), or build each class anchor from a single prompt ($P=1$). Every component contributes on both backbones, and the outer objective has the largest effect: replacing it with MSE lowers accuracy from 92.35\% to 90.13\% on \qwenname{} and from 90.89\% to 88.54\% on \deepseekname{}. Removing the augmented branch lowers accuracy by 1.38 and 1.01 points, the alignment term by 0.73 and 1.07 points, the prompt ensemble by 0.57 and 0.89 points, and the margin by 0.32 and 0.38 points. The alignment term and the prompt ensemble contribute more on \deepseekname{}, whereas the augmented branch contributes more on \qwenname{}.

\begin{table}[t]
    \centering
    \small
    \setlength{\tabcolsep}{3.5pt}
    \caption{Sensitivity to $B_{\mathrm{cls}}$, $\tau$, and $\lambda$ on ImageNet-100 with \qwenname{} at IPC${}=1$. Bold marks the default.}
    \label{tab:hyperparameter_sensitivity}
    \begin{tabular}{cc|cc|cc}
        \toprule
        $B_{\mathrm{cls}}$ & Acc. & $\tau$ & Acc. & $\lambda$ & Acc. \\
        \midrule
        1 & 91.98\pmstd{0.1} & 0.01 & 89.13\pmstd{0.3} & 0.05 & 92.07\pmstd{0.1} \\
        \textbf{2} & \textbf{92.35}\pmstd{0.1} & 0.05 & 91.87\pmstd{0.2} & 0.07 & 92.27\pmstd{0.1} \\
        4 & 92.35\pmstd{0.2} & \textbf{0.07} & \textbf{92.35}\pmstd{0.1} & \textbf{0.10} & \textbf{92.35}\pmstd{0.1} \\
        8 & 92.35\pmstd{0.2} & 0.10 & 92.02\pmstd{0.1} & 0.20 & 90.97\pmstd{0.2} \\
        \bottomrule
    \end{tabular}
\end{table}

\paragraph{Hyperparameter sensitivity.}
Table~\ref{tab:hyperparameter_sensitivity} varies the number of real queries per class $B_{\mathrm{cls}}$, the temperature $\tau$, and the ridge coefficient $\lambda$ on \qwenname{}, one at a time. Accuracy saturates at $B_{\mathrm{cls}}=2$: a single query per class is 0.37 points lower, and larger batches give no further gain. For the temperature, $\tau=0.07$ is best. Setting $\tau=0.01$ makes the softmax overly sharp and lowers accuracy by 3.22 points, while $\tau=0.05$ and $\tau=0.10$ remain within 0.48 points of the best value. For the ridge coefficient, accuracy varies by 0.28 points over $\lambda\in[0.05,0.10]$ and drops by 1.38 points at $\lambda=0.20$, where the adapter is over-regularized.

\FloatBarrier
\section{Conclusion}
\label{sec:conclusion}

We introduced \method{}, which extends dataset distillation for pre-trained models from vision encoders to large generative vision--language models through a token-pooled interface, language--label mixing, and an exact class-space reformulation that reduces the role of the language model to a class kernel. At one image per class, \method{} is within 2.41 points of full-data training on ImageNet-100 and 6.34 points on ImageNet-1K, about half the gap of prior methods on ViT-B encoders. On the same VLMs, it is more accurate than gradient matching on the encoder parameters and up to $21.5\times$ faster, and at five images per class it is within 0.21 points of full-data training on ImageNet-100 with \qwenname{}. Its images transfer to unseen encoders better than those of matching objectives, though less well than real images, and closing this gap is left for future work, together with larger models and larger budgets.

{\small
\bibliographystyle{ieeenat_fullname}
\bibliography{main}
}

\clearpage
\appendix
\section*{Appendix}
\numberwithin{equation}{section}
\numberwithin{table}{section}
\numberwithin{figure}{section}

\section{Derivations}
\label{app:derivations}

This section derives the closed-form adapter, the properties of the mixed targets, the class-space equivalence, and the gradient with respect to the synthetic features.

\subsection{Notation}
\label{app:notation}

Rows index samples. $V_s\in\mathbb{R}^{n\times d_v}$ and $V_r\in\mathbb{R}^{N_q\times d_v}$ are the frozen visual features of the $n$ synthetic images and of $N_q$ real queries. $H\in\{0,1\}^{n\times C}$ is the one-hot label matrix of the synthetic set, with $H_{jc}=1$ if image $j$ has label $c$. $Z\in\mathbb{R}^{C\times d_z}$ contains the mixed targets with rows $z_c$, $G=ZZ^\top$ is the class kernel, and $d_z=d_t+C$.

\subsection{Closed-form adapter}
\label{app:ridge}

The objective in Eq.~\eqref{eq:ridge_objective} is strongly convex for $\lambda>0$. Setting its gradient to zero gives
\begin{equation}
    (V_s^\top V_s+\lambda I_{d_v})\,W = V_s^\top HZ ,
\end{equation}
whose coefficient matrix is positive definite, so the minimizer is unique. The push-through identity
\begin{equation}
    (V_s^\top V_s+\lambda I_{d_v})^{-1}V_s^\top
    = V_s^\top (V_sV_s^\top+\lambda I_n)^{-1}
\end{equation}
yields the sample-space form used in the main paper,
\begin{equation}
    W^\ast = V_s^\top R^{-1}HZ,
    \qquad R = V_sV_s^\top+\lambda I_n .
    \label{eq:app_wstar}
\end{equation}
Both systems are nonsingular even when $V_s$ is rank deficient. In practice we solve with a Cholesky factorization of $R$ and do not form $R^{-1}$.

\subsection{Properties of the mixed targets}
\label{app:targets}

Let the rows $t_c$ of $T$ have unit norm and $Z=[\sqrt{1-\eta}\,T,\ \sqrt{\eta}\,I_C]$ with $0\le\eta\le 1$. The two blocks occupy disjoint coordinates, hence
\begin{align}
    \|z_c\|_2^2 &= (1-\eta)\|t_c\|_2^2+\eta = 1, \\
    G &= ZZ^\top = (1-\eta)\,TT^\top+\eta I_C .
\end{align}
For $i\neq j$,
\begin{equation}
    \|z_i-z_j\|_2^2 = (1-\eta)\|t_i-t_j\|_2^2+2\eta \;\ge\; 2\eta ,
\end{equation}
so a positive $\eta$ separates distinct classes even when their text anchors coincide. If $\mu_j\ge 0$ are the eigenvalues of $TT^\top$, the identity commutes with $TT^\top$ and the eigenvalues of $G$ are $\eta+(1-\eta)\mu_j$. Therefore $\lambda_{\min}(G)\ge\eta$. Since $\operatorname{tr}(TT^\top)=C$ implies $\mu_{\max}\le C$, for $\eta>0$
\begin{equation}
    \kappa_2(G)\le\frac{\eta+(1-\eta)C}{\eta}
    = 1+\frac{C(1-\eta)}{\eta}.
\end{equation}
For fixed $V_s$, the matrix $R$ does not depend on $\eta$, so the choice of targets leaves the conditioning of the visual ridge system unchanged.

\subsection{Proof of Proposition~\ref{prop:class_space}}
\label{app:proof}

By Eq.~\eqref{eq:app_wstar}, the predictions on the queries are
\begin{equation}
    V_rW^\ast = V_rV_s^\top R^{-1}HZ = BZ,
\end{equation}
with $RA=H$ and $B=V_rV_s^\top A$. For a row $b_i$ of $B$, the prediction is $b_iZ$. Its inner products with the targets and its squared norm are
\begin{equation}
    (b_iZ)Z^\top = b_iG,
    \qquad \|b_iZ\|_2^2 = b_iGb_i^\top .
\end{equation}
Since every target has unit norm, the cosine scores are
\begin{equation}
    s_i = \frac{b_iG}{\max\!\big(\sqrt{b_iGb_i^\top},\,\epsilon\big)} ,
\end{equation}
where $\epsilon>0$ is the numerical safeguard of the normalization. The scores, and therefore any loss of the scores and the labels, depend on $Z$ only through $G$. The same holds for the gradients wherever the loss is differentiable. As a corollary, the scores are invariant to any orthogonal transformation $Z\mapsto ZO$ with $OO^\top=I_{d_z}$, since it leaves $G$ unchanged.

\paragraph{Adaptation of another encoder.}
The derivation uses no property of the features other than their role in $R$ and $B$. For a frozen encoder $\phi'$ with features $V_s'=\phi'(\mathcal{S})$ and $V_r'$ of dimension $d_v'$, the adapter fitted to the same targets yields the scores of Eq.~\eqref{eq:class_space_scores} with $R'=V_s'V_s'^\top+\lambda I_n$, $R'A'=H$, and $B'=V_r'V_s'^\top A'$. Adapting $\phi'$ therefore requires the images and the kernel $G$, and neither the targets $Z$ nor a text tower for $\phi'$.

\subsection{Gradient with respect to the synthetic features}
\label{app:adjoint}

\paragraph{Loss and score gradients.}
For one branch, let $\ell_{i,c}=(s_{i,c}-m\,\mathbf{1}[c=y_i])/\tau$ and $p_i=\operatorname{softmax}(\ell_i)$. With $\mathcal{L}=\mathcal{L}_{\mathrm{mar}}+\beta\mathcal{L}_{\mathrm{aux}}$,
\begin{equation}
    g_i=\nabla_{s_i^\top}\mathcal{L}
    =\frac{p_i-e_{y_i}}{N_q\tau}-\frac{\beta\,e_{y_i}}{N_q},
\end{equation}
where $e_y$ is a one-hot vector. Write $c_i=b_i^\top$, $u_i=Gc_i$, and $r_i=\sqrt{c_i^\top Gc_i}$, so that $s_i^\top=u_i/r_i$ for $r_i>\epsilon$. Differentiating gives
\begin{equation}
    \nabla_{c_i}\mathcal{L}
    =\Big(\frac{G}{r_i}-\frac{u_iu_i^\top}{r_i^{3}}\Big)g_i ,
\end{equation}
and the Jacobian is $G/\epsilon$ when $r_i<\epsilon$. Stacking these rows gives $E=\nabla_B\mathcal{L}\in\mathbb{R}^{N_q\times C}$.

\paragraph{Adjoint through the solve.}
With $B=V_rV_s^\top A$, $RA=H$, and $R=V_sV_s^\top+\lambda I_n$,
\begin{align}
    \mathrm{d}B &= V_r(\mathrm{d}V_s)^\top A+V_rV_s^\top\mathrm{d}A, \\
    \mathrm{d}A &= -R^{-1}(\mathrm{d}R)A, \\
    \mathrm{d}R &= (\mathrm{d}V_s)V_s^\top+V_s(\mathrm{d}V_s)^\top .
\end{align}
The direct term contributes $\langle E,V_r(\mathrm{d}V_s)^\top A\rangle=\langle AE^\top V_r,\mathrm{d}V_s\rangle$. For the indirect term, define $P$ by the adjoint system $RP=V_sV_r^\top E$. Since $R$ is symmetric,
\begin{align}
    \langle E,V_rV_s^\top\mathrm{d}A\rangle
    &=-\langle PA^\top,\mathrm{d}R\rangle \notag\\
    &=-\langle (PA^\top+AP^\top)V_s,\mathrm{d}V_s\rangle .
\end{align}
Collecting both terms gives
\begin{equation}
    \begin{aligned}
        RP &= V_sV_r^\top E, \\
        \nabla_{V_s}\mathcal{L} &= AE^\top V_r-(PA^\top+AP^\top)V_s .
    \end{aligned}
    \label{eq:adjoint_gradient}
\end{equation}
The adjoint system reuses the Cholesky factor of $R$, so the gradient requires one additional solve of the same size. The kernel $G$ couples the errors of related classes through $E$, while $R$ is unaffected by the choice of targets. The gradient is then propagated through the feature normalization, the token pooling, the connector and vision tower, the augmentation of the branch, and the image parameterization. Each branch has its own $R_k$, $A_k$, and $P_k$, and the total-variation term is evaluated once on the clean images.

\subsection{Computational cost}
\label{app:cost}

Per branch, forming $R$ costs $O(n^2d_v)$ and its factorization $O(n^3)$. Solving for the $C$ right-hand sides of $H$ costs $O(n^2C)$, compared with $O(n^2d_z)$ for the $d_z=d_t+C$ right-hand sides of $HZ$. Computing $B$ costs $O(N_qnd_v+N_qnC)$ and scoring $O(N_qC^2)$. The kernel $G$ is formed once at cost $O(C^2d_t)$. The adjoint adds one solve of the same size. These counts exclude the forward and backward passes of the vision tower, which are common to all objectives that optimize pixels.

\section{Algorithm}
\label{app:algorithm}

Algorithm~\ref{alg:llmdd} summarizes \method{}. The class kernel is computed once from the prompts, and each iteration performs one closed-form solve per branch.

\begin{algorithm}[t]
\caption{\method{}}
\label{alg:llmdd}
\small
\algrenewcommand\algorithmicindent{1em}
\begin{algorithmic}[1]
\Require Frozen VLM with token-pooled visual interface $\phi$
  and language model $g_\psi$; class prompts $\{p_{c,p}\}$; synthetic images $\mathcal{S}$ with
  one-hot labels $H$; real training data $\mathcal{D}$; ridge
  coefficient $\lambda$; mixing coefficient $\eta$; margin $m$;
  temperature $\tau$; branch weights $\{\omega_k\}_{k=0}^{K}$;
  loss weights $\beta,\gamma$
\Ensure Optimized synthetic set $\mathcal{S}$
\State Compute class anchors $T$ from the prompts
  using Eq.~\eqref{eq:class_anchors}
\State Form the class kernel:
  $G \gets (1-\eta)\,TT^\top + \eta I_C$
\For{$t = 1,\dots,4{,}000$}
  \State Sample a class-balanced real batch from $\mathcal{D}$;
    $V_r \gets \phi(\cdot)$
  \For{$k = 0,\dots,K$}
    \Comment{$k=0$: clean branch}
    \State Synthetic features:
      $V_s \gets \phi(a_k(\mathcal{S}))$
    \State Closed-form adapter:
      $A_k \gets (V_s V_s^\top + \lambda I_n)^{-1} H$
    \State Class-space predictions:
      $B_k \gets V_r V_s^\top A_k$
    \State Cosine scores $s_i$ from $B_k$ and $G$
      using Eq.~\eqref{eq:class_space_scores}
    \State Compute $\mathcal{L}^{(k)} \gets
      \mathcal{L}_{\mathrm{mar}}^{(k)}
      + \beta\,\mathcal{L}_{\mathrm{aux}}^{(k)}$
      using Eq.~\eqref{eq:margin_loss}
  \EndFor
  \State $\mathcal{L}_{\mathrm{out}} \gets
    \sum_{k=0}^{K}\omega_k \mathcal{L}^{(k)}
    + \gamma\,\mathrm{TV}(\mathcal{S})$
  \State Backpropagate through the solves and $\phi$ to obtain
    $\nabla_{\mathcal{S}}\mathcal{L}_{\mathrm{out}}$
  \State Update the synthetic images $\mathcal{S}$
\EndFor
\State \Return $\mathcal{S}$
\end{algorithmic}
\end{algorithm}

\section{Implementation Details}
\label{app:experiment_settings}

\paragraph{VLM interface.}
We use \texttt{Qwen/Qwen3-VL-4B-Instruct} for \qwenname{} and \texttt{deepseek-ai/deepseek-vl2-tiny} for \deepseekname{}. All pre-trained parameters are frozen. Synthetic images are optimized at $256\times256$ pixels for both models. For \qwenname{}, the image is encoded as a single $256\times256$ grid, which replaces the native dynamic resolution. For \deepseekname{}, the image is resized to $384\times384$ and encoded as a single global view, which replaces dynamic tiling. The visual tokens at the output of the connector are averaged and L2-normalized as in Eq.~\eqref{eq:visual_feature}, and the language decoder is not executed on images. Table~\ref{tab:app_interface} lists the resulting dimensions.

\begin{table}[t]
    \centering
    \small
    \setlength{\tabcolsep}{6pt}
    \caption{Token-pooled interface of the two source models.}
    \label{tab:app_interface}
    \begin{tabular}{lcc}
        \toprule
        & \qwenname{} & \deepseekname{} \\
        \midrule
        Encoder input             & $256\times256$ & $384\times384$ \\
        Visual tokens $M$         & 64    & 196 \\
        Feature dimension $d_v$   & 2560    & 1280 \\
        Anchor dimension $d_t$    & 2560    & 1280 \\
        Language decoder on images & not executed & not executed \\
        \bottomrule
    \end{tabular}
\end{table}

\paragraph{Language anchors and targets.}
Each class is described by $P=4$ prompts built from the class name with fixed templates. The prompts are encoded by the language model of the same VLM. For each prompt, the final-layer hidden states are mean-pooled over all non-padding tokens. The prompt representations are then averaged within each class, centered across classes, and L2-normalized as in Eq.~\eqref{eq:class_anchors}. With $\eta=0.25$, the mixed target of class $c$ is $z_c=[\sqrt{0.75}\,t_c,\ 0.5\,e_c]$. The prompts contain no information from evaluation images.

\paragraph{Optimization.}
Synthetic images use the multiscale pyramid parameterization with color decorrelation~\cite{cazenavette2025lgm,olah2017feature} and are updated with Adam~\cite{kingma2015adam} for 4{,}000 steps. The learning rate is 0.05 for the first 3{,}000 steps and then follows a cosine decay to 0.005. We use one clean branch and $K=1$ mildly augmented branch with equal weights $\omega_0=\omega_1=0.5$, and no differentiable Siamese augmentation. Each step samples $B_{\mathrm{cls}}=2$ real queries per class. The margin $m=0.05$ is applied during distillation only, and evaluation uses $m=0$. Table~\ref{tab:app_settings} summarizes the configuration.

\begin{table}[t]
    \centering
    \small
    \setlength{\tabcolsep}{6pt}
    \caption{Default configuration of \method{}.}
    \label{tab:app_settings}
    \begin{tabular}{ll}
        \toprule
        Setting & Value \\
        \midrule
        \qwenname{} checkpoint                  & Qwen3-VL-4B-Instruct \\
        \deepseekname{} checkpoint              & DeepSeek-VL2-Tiny \\
        Image resolution                        & $256\times256$ \\
        Encoder input (\deepseekname{})         & $384\times384$ \\
        Images per class (IPC)                  & 1 \\
        Prompts per class $P$                   & 4 \\
        Mixing coefficient $\eta$               & 0.25 \\
        Ridge coefficient $\lambda$             & 0.10 \\
        Temperature $\tau$                      & 0.07 \\
        Margin $m$ (distillation / evaluation)  & 0.05 / 0 \\
        Alignment weight $\beta$                & 0.2 \\
        TV weight $\gamma$                      & $10^{-5}$ \\
        Augmented branches $K$                  & 1 \\
        Branch weights $\omega_0,\omega_1$      & 0.5, 0.5 \\
        Queries per class $B_{\mathrm{cls}}$    & 2 \\
        Optimizer                               & Adam \\
        Learning rate                           & 0.05, cosine to 0.005 \\
        Image updates                           & 4{,}000 \\
        Independent distillation runs           & 3 \\
        GPU (all but ImageNet-1K)               & RTX A6000, 48\,GB \\
        GPU (ImageNet-1K)                       & NVIDIA H100 \\
        \bottomrule
    \end{tabular}
\end{table}

\section{Baselines and Evaluation Protocol}
\label{app:evaluation}

\paragraph{Real-image baselines.}
Random samples one real training image per class. Centroid selects, for each class, the training image whose feature is nearest to the class mean in the source backbone. Neighbor selects the same-class training image nearest to the final \method{} image in feature space.

\paragraph{Published results.}
The results of LGM and CLP-DD in Table~\ref{tab:prior_work} are taken from~\cite{cazenavette2025lgm} and~\cite{peng2026clpdd}. Both use the ViT-B variants of CLIP, DINOv2, EVA-02, and MoCo-v3 at $224\times224$ resolution and evaluate with a linear probe trained on the distilled images. The LGM results use differentiable augmentation.

\paragraph{Shared-pipeline objectives.}
GM and FM keep the image parameterization, initialization, augmentation branches, real-data sampling, and update budget of \method{} and replace only the objective. GM applies gradient matching~\cite{zhao2021gradient,zhao2021dsa} to the frozen VLM. At every update a 100-way linear head is randomly re-initialized, and for each class the gradients of the cross-entropy loss with respect to the head and to the parameters of the visual encoder and connector are matched between the synthetic and real images. The distance is the cosine distance computed per output unit of each parameter tensor and averaged, and the class losses are averaged. The model weights are not updated, but gradients with respect to them are required, so each update involves second-order backpropagation through the encoder. FM follows distribution matching~\cite{zhao2023distribution}: it minimizes the squared L2 distance between the L2-normalized feature of each synthetic image and the mean L2-normalized feature of the real images of the same class. FM uses no classification head and needs no gradients with respect to the model parameters. Learning rates are selected separately for each objective with a search of equal size on the development split.

\paragraph{Evaluation head.}
All sets distilled or selected on the VLM backbones are evaluated with the same closed-form head. For each method and backbone, we fit the ridge adapter of Eq.~\eqref{eq:ridge_objective} on the frozen features of the images and their labels, using the mixed targets with $\eta=0.25$ and $\lambda=0.10$, and classify the evaluation images by margin-free cosine scores (Eq.~\eqref{eq:class_space_scores} with $m=0$). No network is trained for evaluation.

\paragraph{Cross-model evaluation.}
For Table~\ref{tab:cross_model}, the images distilled with \qwenname{} are passed through each target encoder with its own preprocessing. The target encoders are the ViT-B variants of CLIP, DINOv2, EVA-02, and MoCo-v3 used in~\cite{cazenavette2025lgm,peng2026clpdd}. The head is fitted on the target features with the class kernel $G$ of the source model, as described in Appendix~\ref{app:proof}. The target encoders are neither trained nor used during distillation. The real-image reference uses random training images at the same budget.

\paragraph{Data splits.}
ImageNet-100 uses the class subset of~\cite{tian2020contrastive} with the ImageNet validation images of these classes for final evaluation. Hyperparameters are selected on a development split held out from the training set. For Spawrious and Waterbirds, we follow the benchmark variants, splits, and evaluation protocol of LGM~\cite{cazenavette2025lgm}, and class prompts exclude background labels.

\begin{table}[t]
    \centering
    \small
    \setlength{\tabcolsep}{3.5pt}
    \caption{Same-backbone comparison on ImageNet-100 at IPC${}=1$ with the closed-form head and with a trained linear probe.}
    \label{tab:app_same_backbone}
    \begin{tabular}{lcccc}
        \toprule
        & \multicolumn{2}{c}{Closed-form head} & \multicolumn{2}{c}{Linear probe} \\
        \cmidrule(lr){2-3}
        \cmidrule(lr){4-5}
        Method & \qwenname{} & \deepseekname{} & \qwenname{} & \deepseekname{} \\
        \midrule
        GM                       & 88.23 & 81.77 & 85.23 & 75.46 \\
        FM                       & 85.97& 82.94  & 82.87 & 79.94 \\
        \method{}                & 92.35 & 90.89 & 90.11 & 87.79 \\
        \bottomrule
    \end{tabular}
\end{table}

\paragraph{Reporting.}
Each entry is the mean and standard deviation over three independently distilled sets. Efficiency is measured after warm-up as described in Section~\ref{sec:efficiency}, and timing covers feature computation, the objective, backpropagation, and the image update.

\section{Additional Analysis}
\label{app:additional}

\begin{table*}[t]
    \centering
    \small
    \setlength{\tabcolsep}{6pt}
    \caption{Background-shift evaluation at IPC${}=1$ with standard deviations over three runs. Avg.\ averages the two backbones.}
    \label{tab:background_main}
    \begin{tabular}{lcccccc}
        \toprule
        & \multicolumn{3}{c}{Spawrious}
        & \multicolumn{3}{c}{Waterbirds} \\
        \cmidrule(lr){2-4}
        \cmidrule(lr){5-7}
        Method
        & \qwenname{} & \deepseekname{} & Avg.
        & \qwenname{} & \deepseekname{} & Avg. \\
        \midrule
        Random
        & 27.73\pmstd{7.1} & 19.87\pmstd{8.7} & 23.80\pmstd{7.9}
        & 65.62\pmstd{9.1} & 73.18\pmstd{5.7} & 69.40\pmstd{7.4} \\
        Centroid
        & 25.37\pmstd{4.1} & 32.07\pmstd{7.4} & 28.72\pmstd{5.7}
        & 65.03\pmstd{6.2} & 60.99\pmstd{5.7} & 63.01\pmstd{5.9} \\
        Neighbor
        & 18.55\pmstd{11.1} & 37.41\pmstd{8.9} & 27.98\pmstd{10.1}
        & 56.08\pmstd{4.7} & 68.00\pmstd{3.6} & 62.04\pmstd{4.2} \\
        \midrule
        GM
        & 32.33\pmstd{6.7} & 42.17\pmstd{4.1} & 37.25\pmstd{5.4}
        & 67.12\pmstd{2.1} & 66.21\pmstd{4.9} & 66.67\pmstd{3.5} \\
        FM
        & 28.78\pmstd{6.7} & 29.62\pmstd{7.1} & 29.20\pmstd{6.9}
        & 65.08\pmstd{3.9} & 64.72\pmstd{2.2} & 64.90\pmstd{3.0} \\
        \method{}
        & \textbf{39.41}\pmstd{3.1} & \textbf{54.38}\pmstd{2.9}
        & \textbf{46.90}\pmstd{3.0}
        & \textbf{77.84}\pmstd{1.9} & \textbf{74.23}\pmstd{2.2}
        & \textbf{76.03}\pmstd{2.0} \\
        \midrule
        Full data
        & 71.46\pmstd{0.1} & 68.00\pmstd{0.0} & 69.73\pmstd{0.1}
        & 92.80\pmstd{0.0} & 91.27\pmstd{0.0} & 92.03\pmstd{0.0} \\
        \bottomrule
    \end{tabular}
\end{table*}

\paragraph{Background shift.}
Table~\ref{tab:background_main} repeats the background-shift evaluation of the main paper with standard deviations. We follow the benchmark variants, splits, and protocol of LGM~\cite{cazenavette2025lgm} for Spawrious~\cite{lynch2023spawrious} and Waterbirds~\cite{sagawa2020distributionally}, and class prompts exclude background labels. On Waterbirds, the gain over GM is larger on \qwenname{} (77.84\% versus 67.12\%) than on \deepseekname{} (74.23\% versus 66.21\%).

\begin{figure}[t]
    \centering
    \includegraphics[width=0.7\linewidth]{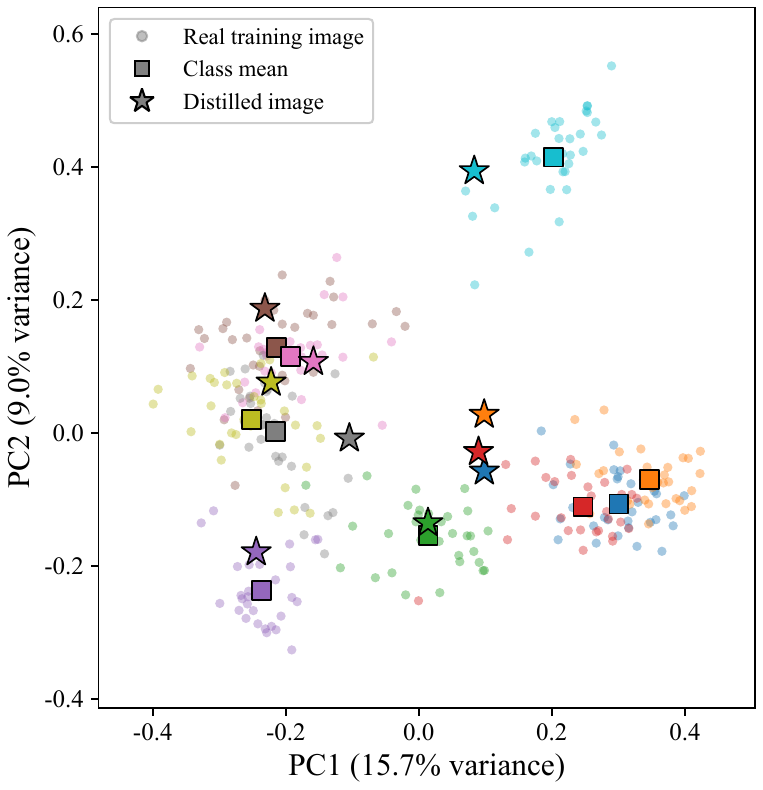}
    \caption{PCA of L2-normalized \qwenname{} features on ImageNet-100 for 10 classes (32 real images per class), fitted on real features only. Stars are the \method{} images at IPC${}=1$.}
    \label{fig:pca}
\end{figure}

\paragraph{Feature-space visualization.}
Figure~\ref{fig:pca} projects \qwenname{} features of real images, class means, and distilled images onto the first two principal components of the real features. Each distilled image lies in or near the region of its own class but does not coincide with the class mean. For robin, little blue heron, and African hunting dog, whose real features overlap, the distilled images are displaced from the overlapping means. The images are thus optimized for the classifier they induce and not for matching class-wise feature statistics. The ten classes are robin, little blue heron, Chesapeake Bay retriever, African hunting dog, ambulance, Dutch oven, mixing bowl, reel, tripod, and stinkhorn. The projection is fitted on 32 real training images per class and explains 24.7\% of the variance with two components.

\begin{figure}[t]
    \centering
    \includegraphics[width=0.75\linewidth]{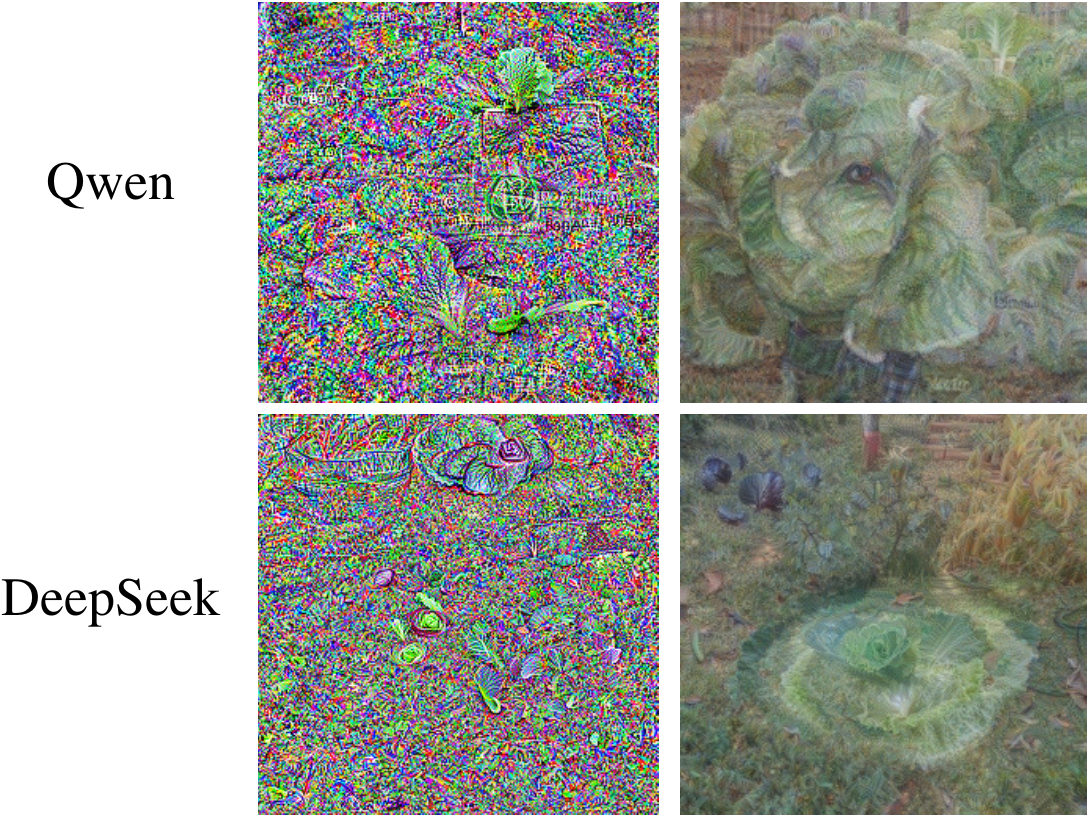}
    \caption{Effect of color decorrelation on ImageNet-100 at IPC${}=1$ (head cabbage) for the \qwenname{} and \deepseekname{} backbones. Left: without color decorrelation. Right: with color decorrelation.}
    \label{fig:color_decorrelation}
\end{figure}

\paragraph{Color decorrelation.}
Following LGM~\cite{cazenavette2025lgm}, \method{} optimizes the synthetic images in a decorrelated color space~\cite{olah2017feature}. Figure~\ref{fig:color_decorrelation} compares images of the same class optimized without and with this component. Without it, the images on both backbones are dominated by saturated high-frequency color noise, whereas with it the same objective produces coherent images in which the object is recognizable. Accuracy is unchanged on both backbones (92.35\% on \qwenname{} and 90.89\% on \deepseekname{}), so the component improves the interpretability of the images at no cost in accuracy on the source model.

\paragraph{Limitations.}
\method{} targets low-IPC adaptation of frozen models. Under background shift, the distilled sets remain well below the full-data reference (Table~\ref{tab:background_main}), so stronger in-distribution adaptation does not by itself provide robustness. The comparison with LGM and CLP-DD in the main paper uses their published results on ViT-B encoders. Images distilled with \method{} transfer to other encoders less well than real images (Table~\ref{tab:cross_model}). The targets depend on the prompt set, and \method{} uses more peak memory than feature-mean matching. All methods are evaluated with the closed-form head, and nonlinear heads, partial fine-tuning, and larger budgets are not evaluated.

\section{Additional Comparisons}
\label{app:additional_comparisons}

\paragraph{Prior methods on VLM backbones.}
Table~\ref{tab:app_same_backbone} adds GM and FM on the token-pooled features of the two VLMs and evaluates all methods with both the closed-form head and a trained linear probe.

\paragraph{Linear gradient matching.}
Table~\ref{tab:app_lgm} reports gradient matching restricted to the linear head, as in LGM~\cite{cazenavette2025lgm}, on the two VLMs.

\begin{table}[t]
    \centering
    \small
    \setlength{\tabcolsep}{3.5pt}
    \caption{Linear gradient matching on the VLM backbones on ImageNet-100 at IPC${}=1$. Time is in seconds per 100 updates on \qwenname{}.}
    \label{tab:app_lgm}
    \begin{tabular}{lccc}
        \toprule
        Method & \qwenname{} & \deepseekname{} & Time \\
        \midrule
        LGM, augmentation of \method{} & 84.22 & 83.91 & 642.17 \\
        LGM, repeated augmentation     & 91.19 & 87.79 & 3218.31 \\
        GM                             & 88.23 & 81.77 & 8784.50 \\
        \method{}                      & 92.35 & 90.89 & 409.37 \\
        \bottomrule
    \end{tabular}
\end{table}

\subsection{Alternative Language-Guidance Mechanisms}
\label{app:language_guidance}
We compare three alternative language-guidance mechanisms for dataset distillation on ImageNet-100 at IPC~$=1$ using Qwen: text-guided token pooling, which aggregates visual tokens using text-derived importance weights; FiLM, which applies text-conditioned channel-wise scaling and shifting to pooled visual features; and cross-attention, which integrates visual and textual information through attention. As shown in Table~\ref{tab:language_guidance}, our method achieves the highest accuracy of 92.35\%, outperforming these alternatives by 35.21, 20.02, and 6.15 percentage points, respectively.

\begin{table}[t]
    \centering
    \small
    \setlength{\tabcolsep}{4pt}
    \caption{Alternative language-guidance mechanisms on ImageNet-100 at IPC${}=1$ using \qwenname{}. Results are Top-1 accuracy (\%), reported as mean $\pm$ standard deviation.}
    \label{tab:language_guidance}
    \begin{tabular}{lc}
        \toprule
        Language-guidance mechanism & Top-1 accuracy \\
        \midrule
        Text-guided token pooling & 57.14\pmstd{1.3} \\
        FiLM                      & 72.33\pmstd{0.2} \\
        Cross-attention           & 86.20\pmstd{0.1} \\
        \method{}                 & \textbf{92.35}\pmstd{0.1} \\
        \bottomrule
    \end{tabular}
\end{table}

\subsection{Bidirectional Transfer between VLM Backbones}
\label{app:vlm_transfer}

We further evaluate whether images distilled using one VLM
backbone remain useful when evaluated using the other backbone.
On ImageNet-100 at IPC~$=1$, we consider both Qwen-to-DeepSeek
and DeepSeek-to-Qwen transfer, where the arrow denotes the
distillation backbone followed by the evaluation backbone.

Table~\ref{tab:vlm_transfer} shows that Qwen-distilled images
achieve 78.12\% accuracy when evaluated with DeepSeek, while
DeepSeek-distilled images achieve 71.83\% when evaluated with
Qwen. The 6.29-percentage-point difference indicates that transfer
performance is asymmetric. Both directions retain classification
utility across backbones, although their accuracies depend on
the source--target pairing.

\begin{table}[t]
    \centering
    \small
    \setlength{\tabcolsep}{4pt}
    \caption{Bidirectional cross-backbone transfer on ImageNet-100 at IPC${}=1$. The arrow indicates the distillation backbone followed by the evaluation backbone. Results are Top-1 accuracy (\%), reported as mean $\pm$ standard deviation.}
    \label{tab:vlm_transfer}
    \begin{tabular}{lc}
        \toprule
        Transfer direction & Top-1 accuracy \\
        \midrule
        \qwenname{} $\rightarrow$ \deepseekname{} & 78.12\pmstd{1.2} \\
        \deepseekname{} $\rightarrow$ \qwenname{} & 71.83\pmstd{1.7} \\
        \bottomrule
    \end{tabular}
\end{table}

\section{Additional Distilled Images}
\label{app:additional_distilled_images}

This section provides additional distilled images for both backbones.

\clearpage

\begin{figure*}[p]
    \centering
    \includegraphics[width=0.85\textwidth,height=0.9\textheight,keepaspectratio]{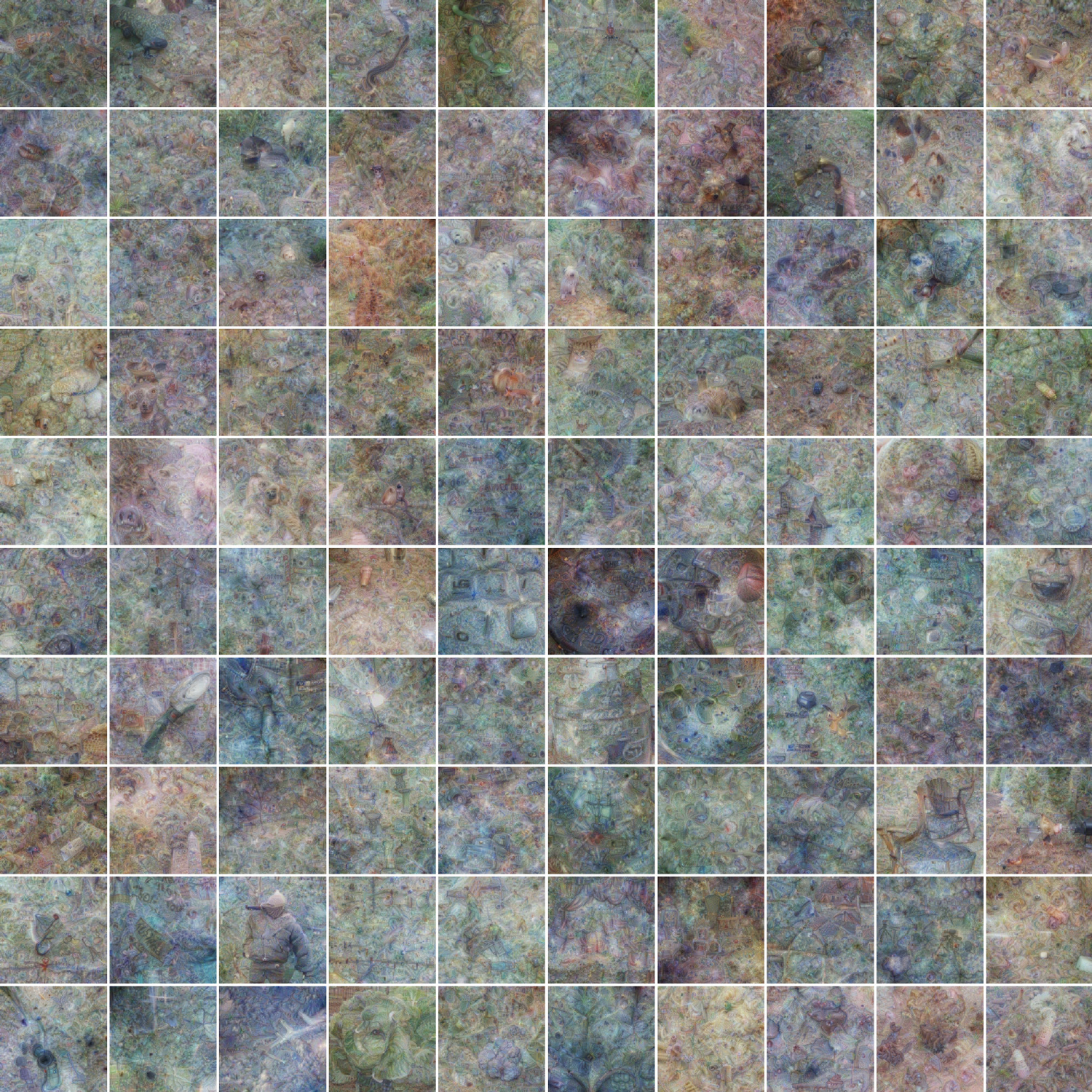}
    \caption{Images distilled by \method{} on ImageNet-100 with the \qwenname{} backbone.}
    \label{fig:app_in100_qwen}
\end{figure*}
\clearpage

\begin{figure*}[p]
    \centering
    \includegraphics[width=0.85\textwidth,height=0.9\textheight,keepaspectratio]{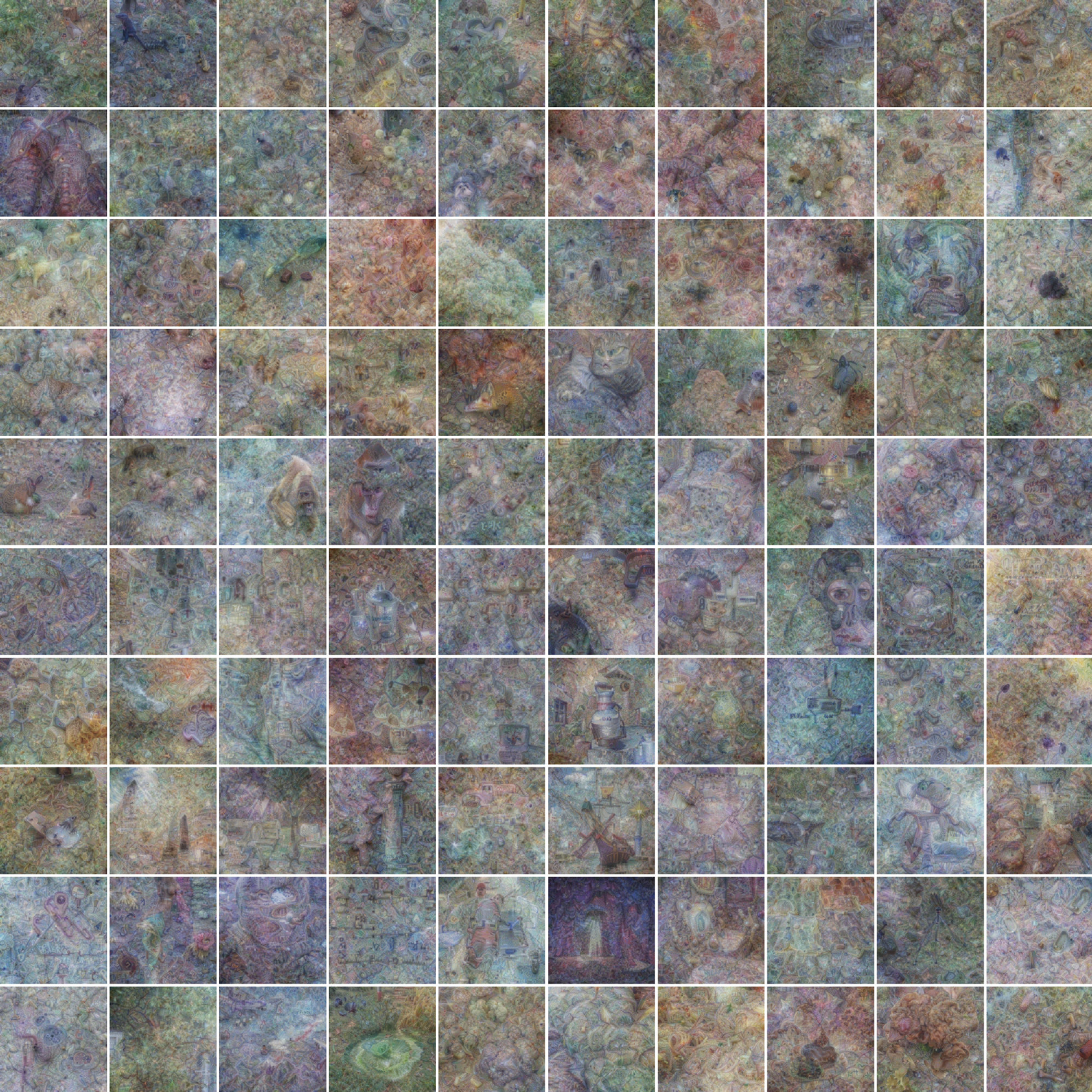}
    \caption{Images distilled by \method{} on ImageNet-100 with the \deepseekname{} backbone.}
    \label{fig:app_in100_deepseek}
\end{figure*}
\clearpage

\begin{figure*}[p]
    \centering
    \includegraphics[width=0.85\textwidth,height=0.9\textheight,keepaspectratio]{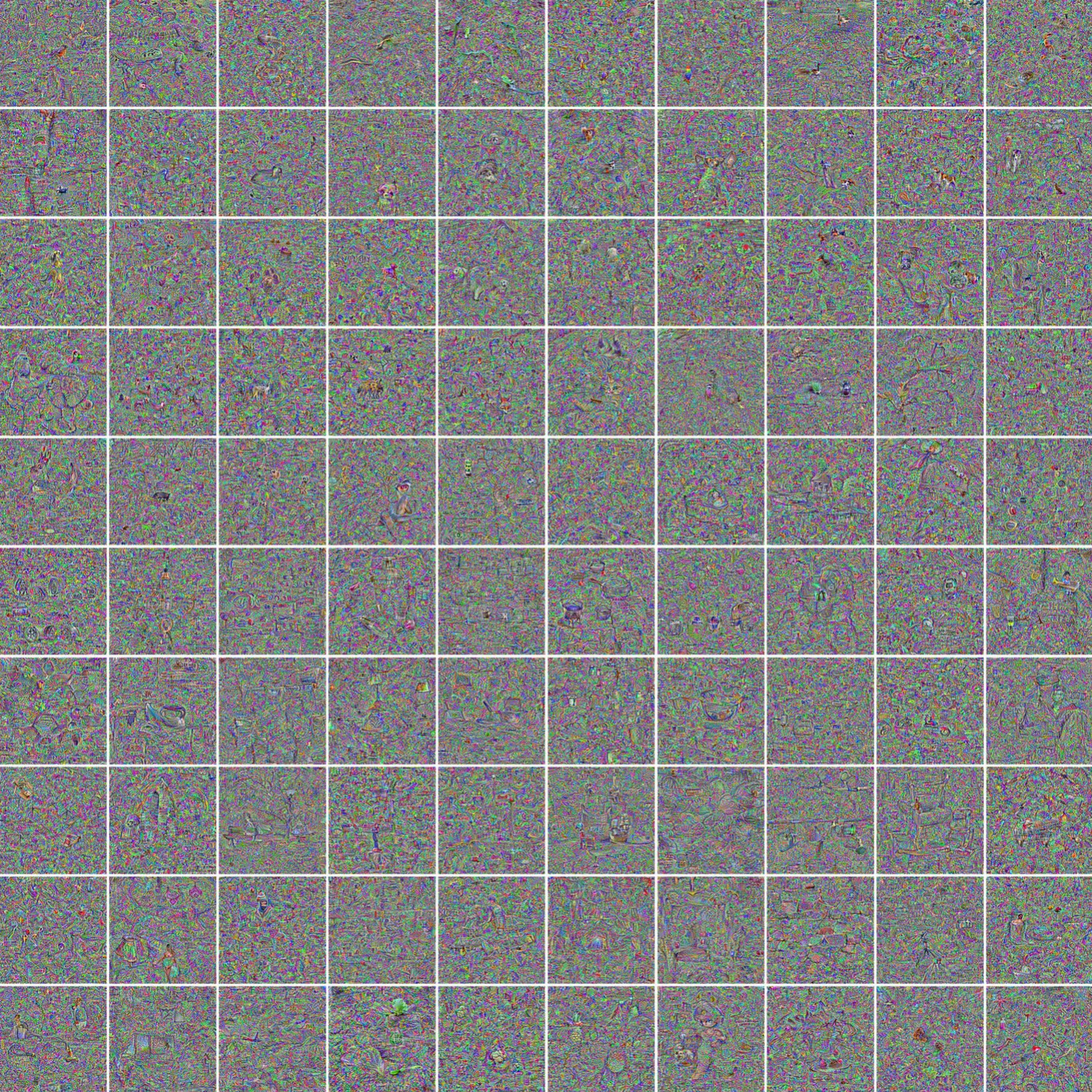}
    \caption{Images distilled by \method{} without color decorrelation on ImageNet-100 with the \qwenname{} backbone.}
    \label{fig:app_in100_qwen_nocolor}
\end{figure*}
\clearpage

\begin{figure*}[p]
    \centering
    \includegraphics[width=0.85\textwidth,height=0.9\textheight,keepaspectratio]{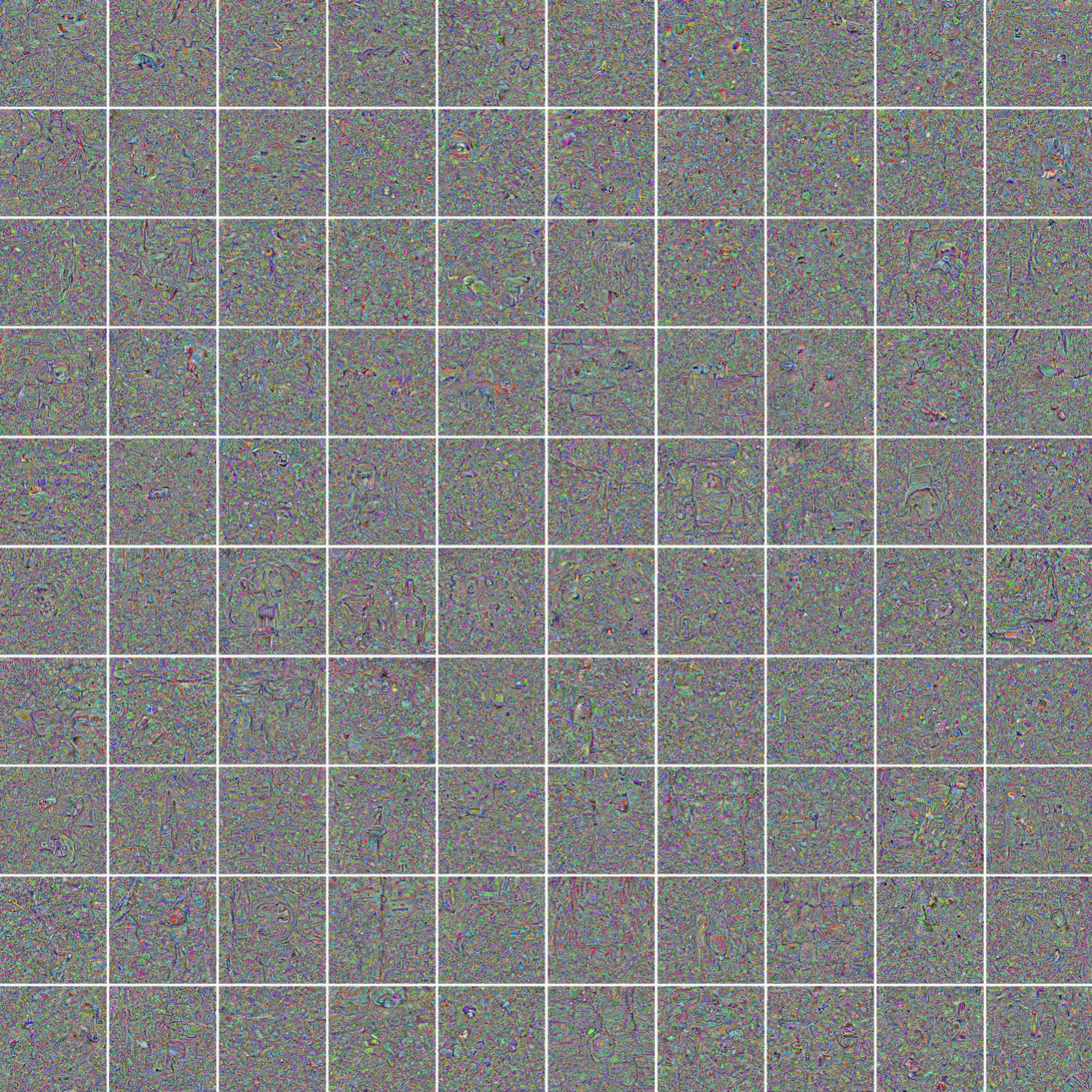}
    \caption{Images distilled by \method{} without color decorrelation on ImageNet-100 with the \deepseekname{} backbone.}
    \label{fig:app_in100_deepseek_nocolor}
\end{figure*}
\clearpage

\begin{figure*}[p]
    \centering
    \includegraphics[width=0.85\textwidth,height=0.9\textheight,keepaspectratio]{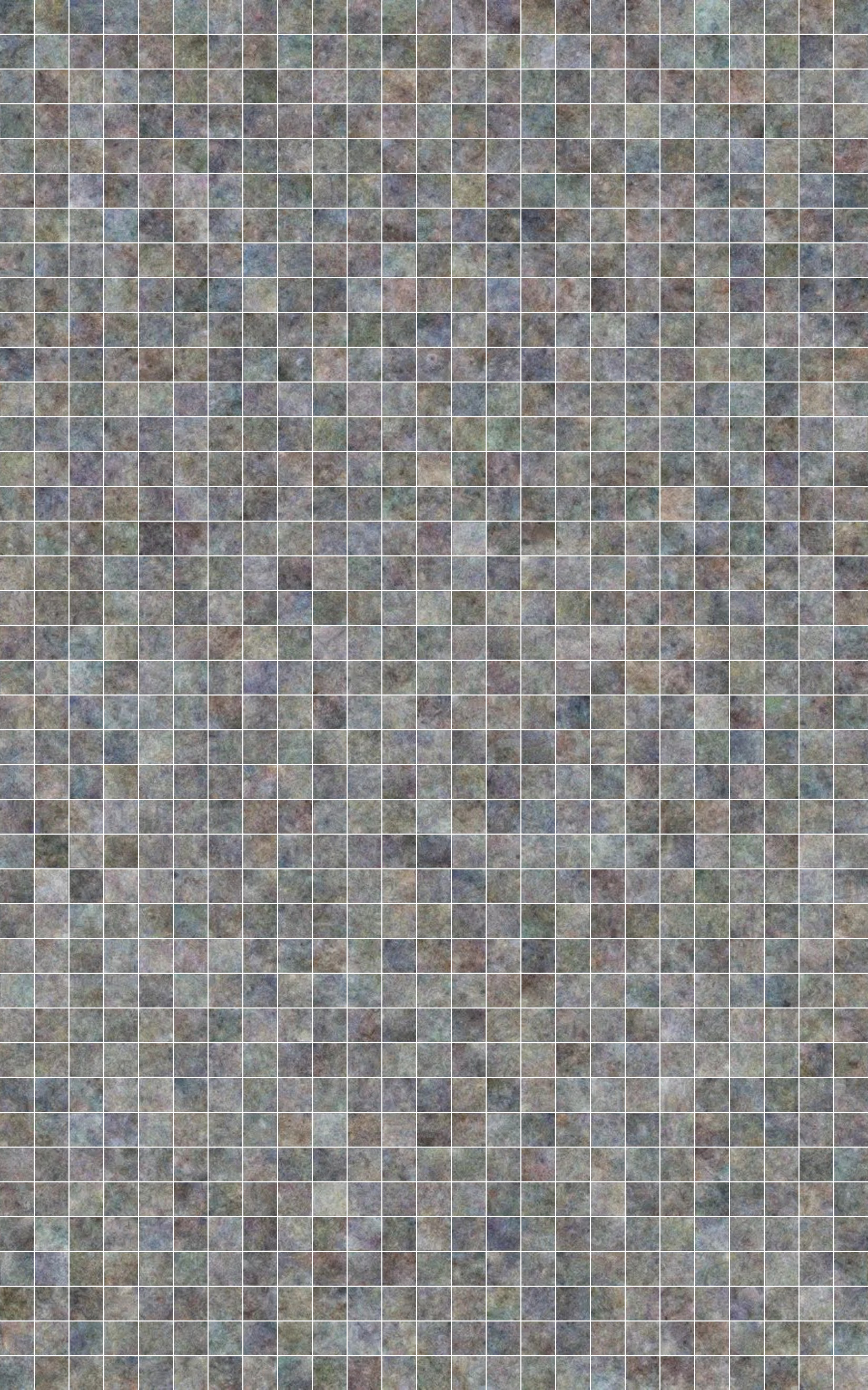}
    \caption{Images distilled by \method{} on ImageNet-1K with the \qwenname{} backbone.}
    \label{fig:app_in1k_qwen}
\end{figure*}
\clearpage

\begin{figure*}[p]
    \centering
    \includegraphics[width=0.85\textwidth,height=0.9\textheight,keepaspectratio]{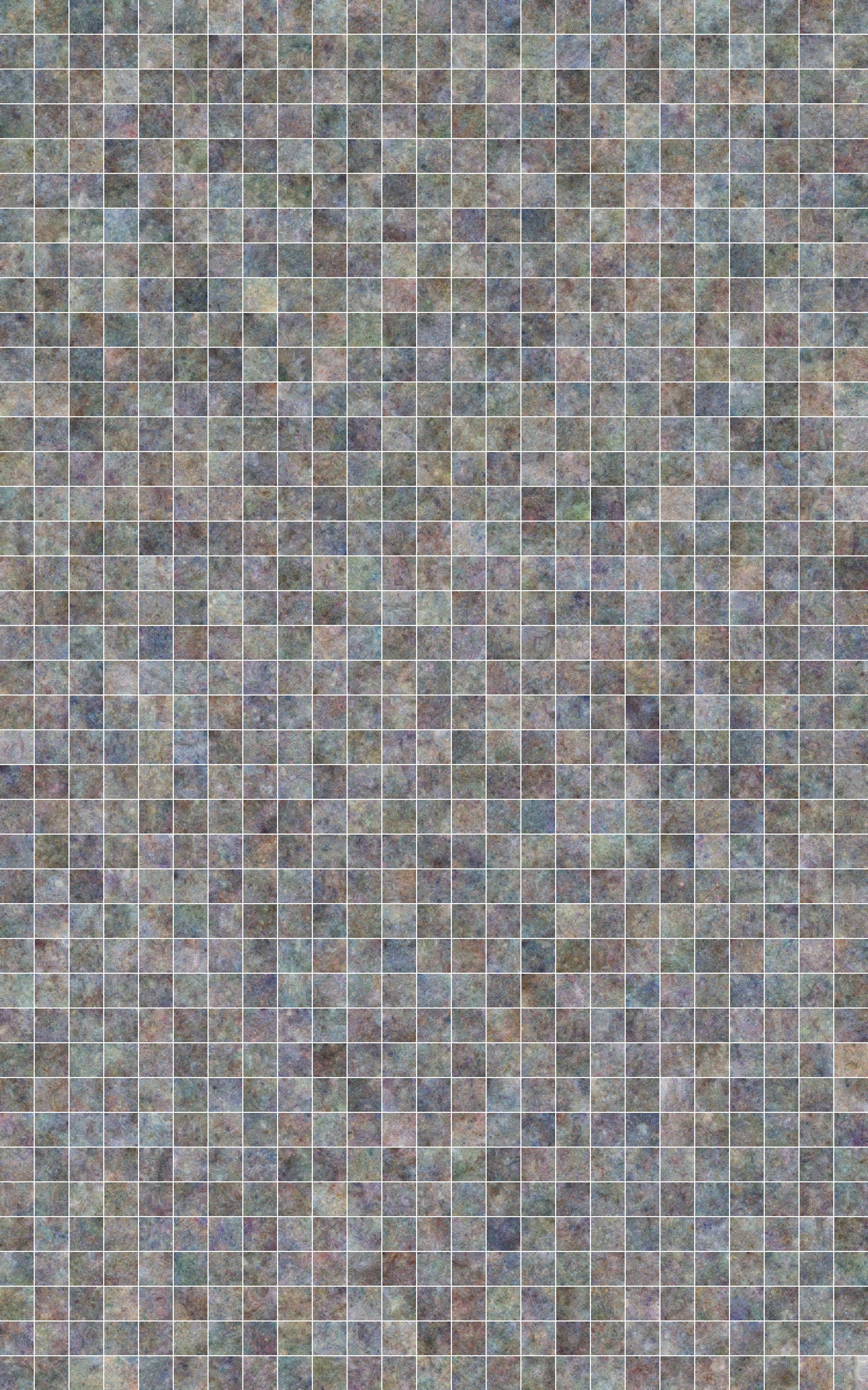}
    \caption{Images distilled by \method{} on ImageNet-1K with the \deepseekname{} backbone.}
    \label{fig:app_in1k_deepseek}
\end{figure*}
\clearpage

\end{document}